\documentclass[journal]{IEEEtran}

\ifCLASSOPTIONcompsoc
\usepackage[nocompress]{cite}
\else
\usepackage{cite}
\fi

\ifCLASSINFOpdf
\usepackage[pdftex]{graphicx}
\graphicspath{{./figures/}}
\DeclareGraphicsExtensions{.jpg,.jpeg,.pdf,.png} 
\else

\fi

\usepackage{dsfont}
\usepackage{amsmath}
\usepackage{amssymb}
\usepackage{bbm}

\usepackage[caption=false,font=footnotesize,labelfont=sf,textfont=sf]{subfig}

\usepackage[usenames, dvipsnames]{color}
\usepackage[dvipsnames]{xcolor}
\usepackage[hidelinks]{hyperref}
\usepackage{enumitem}

\usepackage{soul}

\usepackage{setspace}
\usepackage{tabu} 

\usepackage{times}
\usepackage{epsfig}
\usepackage{graphicx}
\usepackage{amsmath}
\usepackage{amssymb}
\usepackage{xcolor}
\usepackage{array}
\usepackage{multirow}
\usepackage{amsthm,amsmath,amssymb}
\usepackage{mathrsfs}
\usepackage{graphicx}
\usepackage{amsmath}
\usepackage{colortbl} 
\usepackage{amssymb}
\usepackage{booktabs}
\usepackage{times}
\usepackage{microtype}
\usepackage{epsfig}
\usepackage{placeins}
\definecolor{baselinecolor}{gray}{0.9}
\newcommand{\baseline}[1]{\cellcolor{baselinecolor}{#1}}

\newcommand{\revise}[1]{\textcolor{black}{#1}}
\renewcommand{\figurename}{Fig.}

\newif\ifcomments
\commentstrue 

\newcommand{\mr}[1]{\textcolor{black}{#1}}

\usepackage[switch]{lineno}

\begin{document}


\title{Jointly Understand Your Command and Intention:\\ Reciprocal Co-Evolution between Scene-Aware 3D Human Motion Synthesis and Analysis}

\author{Xuehao Gao,
	Yang Yang,
	Shaoyi Du,
	Guo-Jun Qi,
	and Junwei Han
}

\maketitle

	
\begin{abstract}

\mr{As two intimate reciprocal tasks, scene-aware human motion synthesis and analysis require a joint understanding between multiple modalities, including 3D body motions, 3D scenes, and textual descriptions. In this paper, we integrate these two paired processes into a Co-Evolving Synthesis-Analysis (CESA) pipeline and mutually benefit their learning.} Specifically, scene-aware text-to-human synthesis generates diverse indoor motion samples from the same textual description to enrich human-scene interaction intra-class diversity, thus significantly benefiting training a robust human motion analysis system. Reciprocally, human motion analysis would enforce semantic scrutiny on each synthesized motion sample to ensure its semantic consistency with the given textual description, thus improving realistic motion synthesis. Considering that real-world indoor human motions are goal-oriented and path-guided, we propose a cascaded generation strategy that factorizes text-driven scene-specific human motion generation into three stages: goal inferring, path planning, and pose synthesizing. Coupling CESA with this powerful cascaded motion synthesis model, we jointly improve realistic human motion synthesis and robust human motion analysis in 3D scenes.


\end{abstract}


\section{Introduction}
\label{sec:intro}
\IEEEPARstart{S}{y}nthesizing realistic human motions inside 3D scenes from textual descriptions brings broad applications into the real world, including VR content creation, digital animation design, and film script visualization \cite{hu2021text,youwang2022clip,guo2022tm2t,wang2022humanise,jiang2023motiongpt}. Besides, via integrating this text-to-motion generation technique into a humanoid robot platform, a language-based description can also serve as a control signal for instructing an intelligent robot \cite{gong2023arnold,wang2021digital,lee2002interactive,yuan2020residual,wang2019combining,zhou2023ude}. Although many recent fruitful attempts have been made to generate realistic human motions from textual descriptions ($e.g.$, \textit{``a person stands up and then runs in a circle"}), most of them synthesize body motions in isolation from the environmental context, thus leaving human-scene interactions behind \cite{chen2023executing,tevet2022human,gao2023decompose,petrovich2022temos,yuan2022physdiff,dabral2023mofusion}. However, real-world human movements are goal-directed and influenced by the spatial layout of their surrounding scenes ($e.g.$, \textit{walk to an armchair and sit down}) \cite{cao2020long,wang2021synthesizing,hassan2021stochastic,huang2023diffusion}. Therefore, to synthesize realistic human-scene interactions, this paper reformulates the conditional human motion generation task into a cross-modal joint inference problem based on the given textual command and 3D scene context.

\begin{figure}[t]
	\centering
	\includegraphics[width=0.49\textwidth]{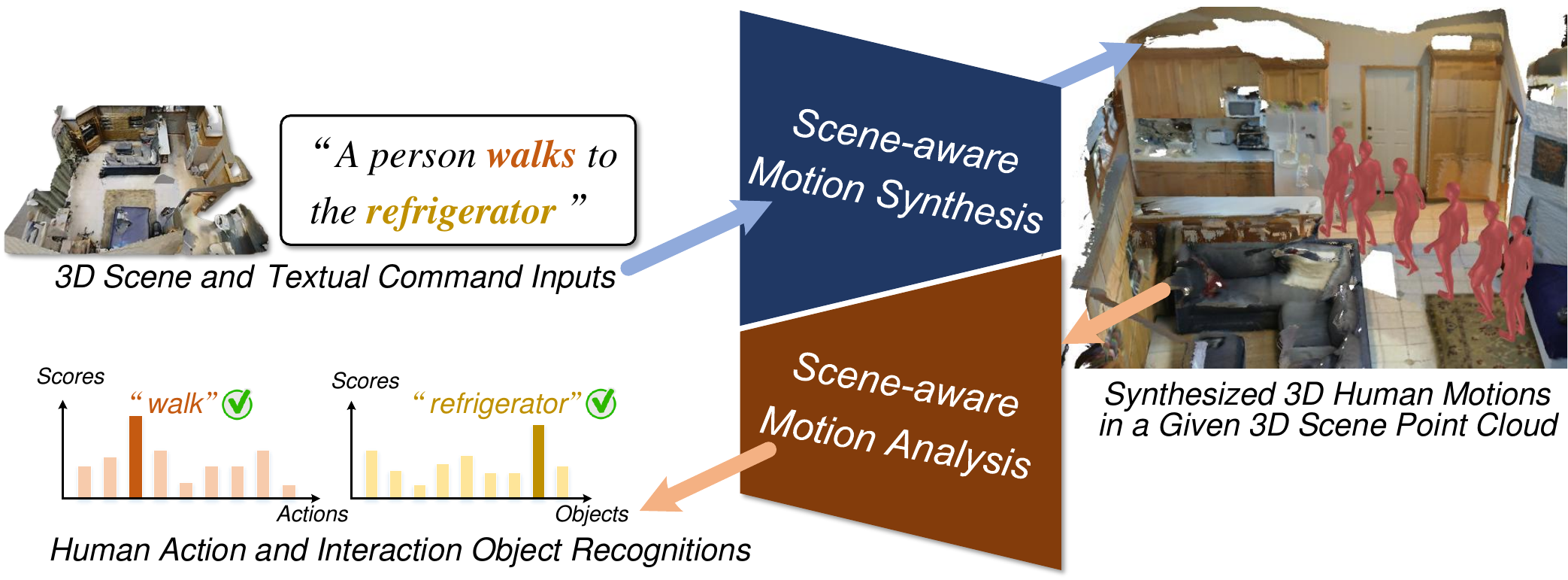}	
	\vspace{-4mm}
	\caption{Co-evolving Synthesis-Analysis Pipeline. Scene-aware text-to-motion generation synthesizes 3D indoor human poses conditioned on the given text commands and 3D scenes. Given a synthesized human-scene interaction sample, scene-aware motion analysis recognizes its human action and interaction object categories.}
	\label{Pipeline}
\end{figure}

As shown in Fig. \ref{Pipeline}, the bi-directional semantic mapping between vision and language bridges scene-aware human motion generation and analysis as two intimate reciprocal tasks \cite{vazquez2018words,yang2023antifreezing}. As a reverse process of scene-aware text-to-motion generation \cite{gao2021efficient}, human motion analysis focuses on inferring the semantics of an observed 3D indoor human motion, including recognizing its activity category and the object it interacts with \cite{aggarwal1999human,hassan2019resolving,zhang2021learning,huang2022capturing}. Therefore, human motion analysis enables an intelligent machine to understand human behaviors and analyze one's intentions for planning its own reactions \cite{guo20143d,gupta2015indoor,aggarwal2014human,liu2019ntu}. 

 
The core insight missing is that integrating scene-aware motion synthesis and analysis into a synergistic pipeline would benefit their learning from each other: (I) \textit{Diverse generation encourages robust recognition.} Given the same textual description and 3D scene, non-deterministic text-to-motion synthesis generates diverse human-scene interaction samples with different motion styles, speeds, paths, \textit{etc}. These synthesized diverse motion samples significantly enrich the intra-class diversity of human-scene interactions and thus benefit training a robust scene-aware human motion analysis system; (II) \textit{Powerful recognition fosters realistic generation.} Human motion analysis can act as a post-discriminator and scrutinize the semantics of synthesized human-scene interactions by recognizing their activity classes and interaction objects, thus ensuring text-motion consistency. These synergistic benefits between scene-aware motion synthesis and analysis tasks inspire us to integrate them into a synergistic pipeline ($i.e.$, CESA). 


Notably, text-conditioned human motion synthesis and analysis in 3D scenes are two quite challenging multi-modal understanding tasks that both require a holistic and joint understanding of scenes, motion, and language. Specifically, scene-specific text-to-motion generation focuses on learning a cross-modal inference from a text-scene pair to motions. Considering that real-world indoor human motions are goal-oriented and path-guided, we propose a cascaded generation strategy that factorizes the challenging scene-aware text-to-motion synthesis task into three relatively easy stages: given a 3D scene and textual command, we first infer a motion goal inside this scene, then plan a movement path towards this goal, and finally synthesize 3D poses following this path. Furthermore, to facilitate cross-modal semantic inference in the scene-aware human motion analysis task, we develop a powerful human-scene interaction recognizer that jointly understands an indoor human motion and its environmental contexts and infers the action semantics and the interacted object. Coupling these two powerful scene-aware human motion synthesis and analysis systems into a co-evolutionary synthesis-analysis pipeline CESA, we improve the holistic and joint understanding of scenes, motion, and language, significantly benefiting these two motion-related reciprocal tasks.


The core contributions of this paper are summarized as: 
\begin{itemize}
\item We propose a novel multi-modal inference system named CESA that integrates scene-aware human motion synthesis and analysis into a synergistic pipeline and collaboratively improves these two motion-related reciprocal tasks; 
\item We propose a powerful scene-aware text-to-motion model that effectively synthesizes goal-oriented and path-guided 3D human motions inside 3D scenes from their textual descriptions;
\item \revise{Coupling the scene-aware text-to-motion model with CESA, we develop a high-performance human motion synthesis system that outperforms state-of-the-art methods by a large margin in motion realism, text-motion consistency, and motion-scene compatibility.}
\end{itemize}

\section{Related Work}
\label{related_work}
\subsection{Text-Conditioned Human Motion Synthesis}
\label{t2m}
\revise{Synthesizing realistic human motions inside 3D scenes from given text-based descriptions brings broad applications into the real world, including VR content creation, digital animation design, and film script visualization \cite{wang2019combining, aristidou2022rhythm,hou2023two,loi2023machine,boukhayma2018surface,yan2018structure}. Recent human motion synthesis methods can be grouped into different specific generative sub-tasks based on their different condition input types, such as start-end positions \cite{wang2021synthesizing, hassan2021stochastic}, movement trajectories \cite{wang2024pacer+, karunratanakul2023guided}, scene contexts \cite{cao2020long, wang2022humanise}, textual descriptions \cite{gao2024guess, jiang2023motiongpt}, background musics \cite{li2021ai, zhuang2022music2dance}, and speech audios \cite{liu2022beat, qian2021speech}. These diverse modalities of condition inputs reflect special requirements for human motion synthesis in different application contexts.}

Intuitively, text-driven human motion synthesis can be viewed as a text-to-motion translation task. The inherent many-to-many mapping problem behind this task makes synthesizing realistic and diverse 3D human motions very challenging \cite{kim2023flame,lin2023being}. For example, the same word ``walking" can refer to diverse human walking samples with different styles, paths, and speeds \cite{cervantes2022implicit,tang2023flag3d,arkushin2023ham2pose}. Meanwhile, we can also describe the same human motion sample with different words and grammatical forms. \revise{For example, MLD \cite{chen2023executing}, Mofusion \cite{dabral2023mofusion}, and GUESS \cite{gao2024guess} tend to develop a diffusion-based generative scheme for the text-to-motion synthesis task. However, diffusion-based generative scheme also have some potential limitations, such as slower generation process compared to some other generative models, limited applicability to certain data types (e.g., structured skeleton data), lower likelihood compared to other models, and an inability to perform dimensionality reduction \cite{hao2022learning, hao2022cdfkd, haodata, hao2024one}.} Furthermore, most of recent text-to-motion synthesis works solely focus on the text-specific condition and generate body motions in isolation from the 3D scene context, thus leaving the human-scene interaction behind \cite{chen2023executing,tevet2022human,petrovich2022temos,tevet2022motionclip,yuan2022physdiff,dabral2023mofusion}. As an under-explored task, text-driven human motion generation in 3D scenes incorporates the joint conditional contexts of scenes and textual descriptions into synthesizing human-scene interaction samples, thus enriching diverse motion intentions \cite{wang2022humanise}. Compared with the conventional text-to-motion synthesis \cite{wang2021synthesizing,wang2021scene,zhao2022compositional,araujo2023circle}, generating human motions in 3D scenes from their textual descriptions is much more challenging as: (1) its synthesized human motions are jointly conditioned on the multi-modality prior, including 3D scene layout and language-based description; (2) its synthesized human motions should be physically plausible and contextually compatible inside 3D scenes. 

\subsection{Human Motion Semantics Understanding}
Inferring semantic labels from given human motion sequences is the core task of action recognition and is also fundamental for various real-world applications, including autonomous driving \cite{martin2019drive,li2024ggrt,li2024gp}, intelligent surveillance \cite{khan2020human,li2024dgtr}, and human-machine interaction \cite{rodomagoulakis2016multimodal, xue2024weakly}. In recent years, 3D skeleton/mesh-based action recognition is attracting increasing interests \cite{gao2023glimpse,gao2023learning,gao2021efficient,wang2021dear,hu2019joint}. Compared with RGB videos, 3D-based skeleton or mesh modality is a more well-structured body representation and has better robustness against environmental noises (\textit{e.g.}, observation viewpoint, background clutter, lighting condition, clothing appearance), allowing action recognition algorithms to focus on the robust action-specific features. Among recent action recognition systems, how to develop a powerful feature extractor for effective spatial-temporal movement pattern learning (\textit{i.e.}, intra-frame posture and inter-frame motion modeling) is a challenge that remains under-explored \cite{liu2023skeleton, liu2020disentangling, pmlr-v148-gao21a,yan2018spatial,pang2023skeleton,liu2020multi}.

\subsection{Human-Scene Interaction Analysis}
Human-scene interaction analysis is crucial for an intelligent agent to understand human manipulation intentions inside 3D scenes \cite{gammulle2023continuous,yan2019research,gao2024multi,wang2022happens,sun2023learning}. As a reverse process of text-to-motion generation, scene-aware human motion analysis focuses on inferring core semantic information from an observed human-scene interaction sample, including recognizing its activity category and interaction object \cite{jaouedi2016human,wang2021digital}. Compared with the text-driven ``many-to-many" human motion generation, human motion analysis is a deterministic ``many-to-one" problem in which many different human-scene interaction samples may refer to the same activity semantics or interaction object. Therefore, sufficient labeled human-scene interaction samples are one of the key factors for learning a robust human motion analysis model \cite{jhuang2013towards,idrees2017thumos,zhang2016action}. As two naturally paired tasks, we integrate scene-aware human motion synthesis and analysis into a synergistic pipeline and benefit their learning from each other. As an initial attempt, we hope it will inspire the community for more exploration.

\section{Co-Evolving Synthesis-Analysis Pipeline}
\subsection{Notations}
In this paper, we introduce CESA, which integrates scene-aware human motion synthesis and analysis into a synergistic pipeline and explores reciprocal benefits between them. The training samples used in CESA are text-motion-scene pairs as $\{\boldsymbol{T}, \boldsymbol{M}, \boldsymbol{S}\}$. $\boldsymbol{S} \in \mathbb{R}^{P\times6}$ denotes a RGB-colored 3D scene with $P$ points, whose three dimensions are for positions and the remaining for colors. We represent a $N$-frame human motion sequence inside $\boldsymbol{S}$ with SMPL-X \cite{pavlakos2019expressive} parameters as $\boldsymbol{M}=[\boldsymbol{M}_1, \boldsymbol{M}_2, \cdots, \boldsymbol{M}_N]$, where $\boldsymbol{M}_{n}=\text{SMPL}(\boldsymbol{t}_n, \boldsymbol{r}_n, \boldsymbol{\beta}, \boldsymbol{p}_n)$. At $n$-th frame, $\boldsymbol{t}_n \in\mathbb{R}^{3}$ denotes the global translation, $\boldsymbol{r}_n\in \mathbb{R}^6$ denotes the global orientation, $\boldsymbol{\beta} \in\mathbb{R}^{10}$ is the body shape parameters, and $\boldsymbol{p}_n\in\mathbb{R}^{J\times3}$ denotes $J$-joint rotations along coordinate axis. Thus, we denote generated body pose parameters at $n$-th frame with $\boldsymbol{m}_{n}$ as $\boldsymbol{m}_{n}=[\boldsymbol{t}_n,\boldsymbol{r}_n,\boldsymbol{p}_n]$. Following Sr3D \cite{achlioptas2020referit3d}, the compositional template of textual description $\boldsymbol{T}$ is: $<$\small{ACTION}$><$\small{OBJECT}$><$\small{RELATION}$><$\small{ANCHOR}$>$, annotating $\boldsymbol{M}$ inside $\boldsymbol{S}$, such as \textit{$<$sit on$>$$<$the armchair$>$$<$ near$>$$<$the door$>$}.
\subsection{Overview}
\begin{figure*}[t]
	\centering
	\includegraphics[width=1\textwidth]{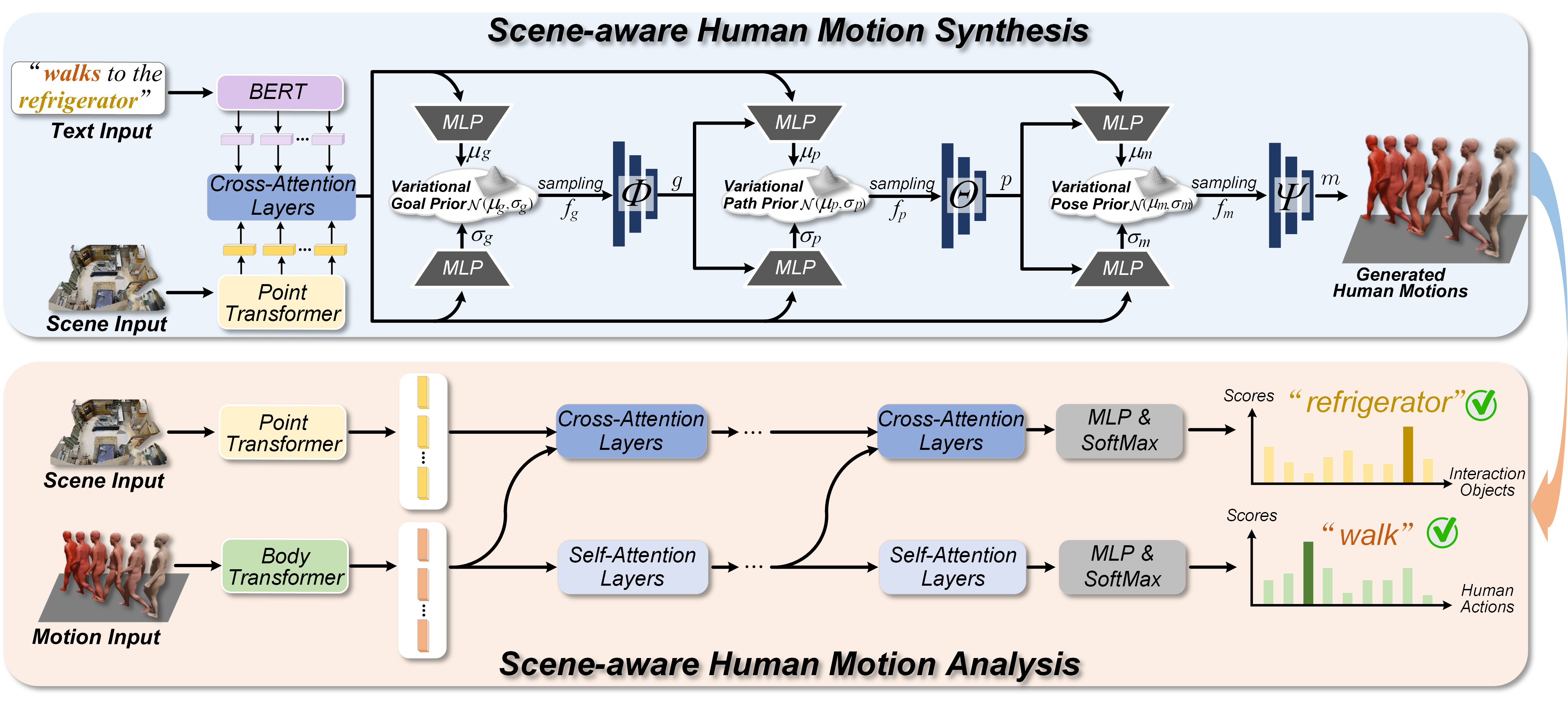}	
	\caption{Core Components. Given encoded text feature $\boldsymbol{f}_{T}$ and scene feature $\boldsymbol{f}_{S}$, text-to-motion generator parameterizes a set of Gaussian latent spaces from $\boldsymbol{f}_{T}$ and $\boldsymbol{f}_{S}$ for non-deterministic path planning and pose synthesizing. Then, the motion analyzer extracts latent features from a synthesized human-scene interaction sample to recognize its action category and interaction object.}
	\label{component}
\end{figure*}

As sketched in Fig. \ref{component}, the core of human motion synthesis lies in formulating a powerful generator $\mathcal{F}_{g}(\cdot)$ for synthesizing realistic and diverse human motions inside 3D scenes from textual instructions as: $\boldsymbol{M}=\mathcal{F}_{g}(\boldsymbol{T},\boldsymbol{S})$. In the human motion analysis, analyzer $\mathcal{F}_{a}(\cdot)$ recognizes the activity category \small{ACT} and interaction object \small{OBJ} from an observed human-scene interaction sample as: $[$\small{ACT}, \small{OBJ}$]$ = $\mathcal{F}_{a}(\boldsymbol{M},\boldsymbol{S})$. In this paper, we integrate these two reciprocal tasks into a co-evolving pipeline and explore synergistic benefits between them. We will elaborate on their technical details in the following sections.

\subsection{Scene-aware Motion Generator}
Since real-world human motions are goal-oriented and path-guided, we propose a cascaded conditional variational autoencoder that factorizes text-driven scene-specific motion synthesis into three sequential stages: goal inferring, path planning, and pose synthesizing.
\subsubsection{\textit{Multi-Modal Encoder}: Model Text-Scene Conditions}
In the multi-modal encoder, we extract latent representations from given scene and text input and integrate them into a joint conditional context embedding. Specifically, given a language-based instruction input $\boldsymbol{T}$, we first deploy a pre-trained BERT \cite{kenton2019bert} as a text feature extractor to encode it into a $P$-token embedding sequence $\boldsymbol{f}_{T}$ as $\boldsymbol{f}_{T}=[\boldsymbol{f}_{T}^{1},\cdots,\boldsymbol{f}_{T}^{P}]$. To model the 3D scene condition input, we employ a pre-trained Point Transformer \cite{zhao2021point} as a scene feature extractor to encode $\boldsymbol{S}$ into its $Q$-token embedding sequence $\boldsymbol{f}_{S}$ as $\boldsymbol{f}_{S}=[\boldsymbol{f}_{S}^{1}, \cdots, \boldsymbol{f}_{S}^{Q}]$. Then, we use a cross-attention-based multi-modal fusion module \cite{vaswani2017attention} to integrate $\boldsymbol{f}_{S}$ and $\boldsymbol{f}_{S}$ into a joint conditional context embedding $\boldsymbol{f}_{ST}$ as: $\boldsymbol{f}_{ST} =$\textit{\ttfamily{CrossAtt}}$(\boldsymbol{f}_{S},\boldsymbol{f}_{T})$. 

\subsubsection{\textit{Goal Decoder}: Infer Destination from Instruction} 
\begin{figure}[t]
	\centering
	\includegraphics[width=0.5\textwidth]{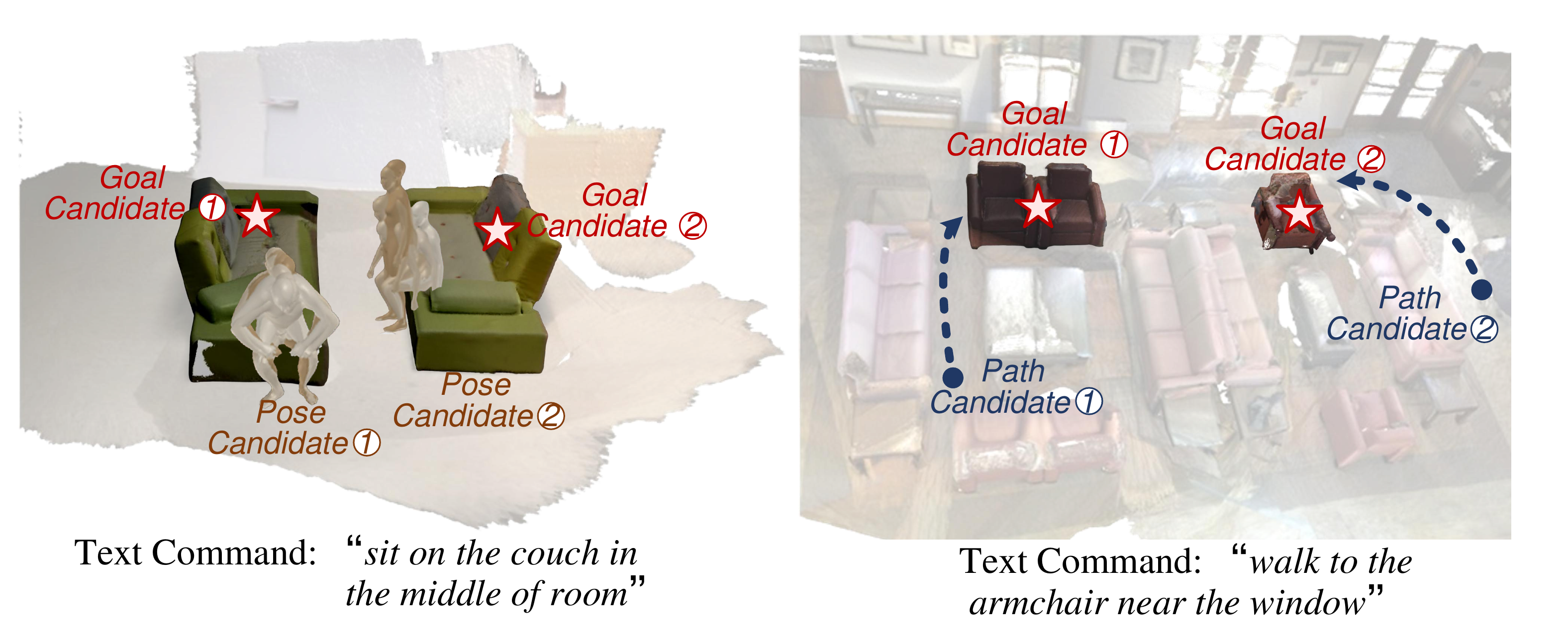}	
	\vspace{-4mm}
	\caption{Non-one-on-one corresponding in text-to-motion synthesis in 3D scenes. In some scene contexts, more than one indoor goal/path/pose may both conform to the description in the given textual command.}
	\label{multiple_goals}
\end{figure}

Synthesizing realistic human motions in a 3D scene begins with inferring the correct movement destination from a textual command. Notably, suffering from the ambiguity in linguistic descriptions, there may be more than one indoor object in a given scene context that concurrently conforms to a textual description for the intended target. For example, as shown in Fig. \ref{multiple_goals}, given a textual command as “\textit{walk to the armchair near the window}", two \textit{armchair} candidates in the scene both conform to this description. To this end, we develop a probabilistic Goal Decoder for non-deterministic movement destination inference. 

Specifically, we consider the 3D center position of the object to interact with (\textit{i.e.}, OBJ) as the motion goal $\boldsymbol{g}\in\mathbb{R}^{3}$. Given scene-text joint condition embedding $\boldsymbol{f}_{ST}$, goal decoder $\mathit{\Phi}$ infers a non-deterministic movement goal $\boldsymbol{g}$ inside a specified scene via exploring the uncertainty behind the posterior probability distribution $\mathcal{P}(\boldsymbol{g}|\boldsymbol{f}_{ST})$ as:
\begin{equation}
	\begin{aligned}
		\boldsymbol{f}_g \sim & \mathcal{Q}(\mathcal{Z}_{g}|\boldsymbol{f}_{ST}) \equiv \mathcal{N}(\boldsymbol{\mu}_{g}, \boldsymbol{\sigma}_{g}), \\
		\textit{where} \; \boldsymbol{\mu}_{g} & = \textit{\ttfamily{MLP}}_{g}^{1}(\boldsymbol{f}_{ST}), \boldsymbol{\sigma}_{g} = \textit{\ttfamily{MLP}}_{g}^{2}(\boldsymbol{f}_{ST}). 
	\end{aligned}
	\label{Eq.0}
\end{equation}
As shown in Eq.\ref{Eq.0}, we first deploy two \textit{\ttfamily{MLP}} layers to map $\boldsymbol{f}_{ST}$ into two Gaussian distribution parameters $\boldsymbol{\mu}_{g}$ and $\boldsymbol{\sigma}_{g}$. Then, we parameterize a latent goal feature space $\mathcal{Z}_{g}$ based on $\boldsymbol{\mu}_{g}$ and $\boldsymbol{\sigma}_{g}$. In this way, we estimate a variational goal posterior $\mathcal{Q}$ by assuming a Gaussian information bottleneck. Finally, we sample a latent-based goal feature $\boldsymbol{f}_{g}$ from $\mathcal{Z}_{g}$ and send it into a Transformer-based goal decoder $\mathit{\Phi}$ for decoding a inferred 3D body motion $\boldsymbol{\overline{g}}$ as:
\begin{equation}
	\boldsymbol{\overline{g}} = \mathit{\Phi}(\boldsymbol{f}_{g}).
\end{equation}
In the training stage, we optimize goal decoder $\mathit{\Phi}$ by minimizing two objectives: (1) the $l_1$-based distance error between the predicted 3D goal $\boldsymbol{\overline{g}}$ and its ground-truth $\boldsymbol{g}$; (2) the Kullback-Leibler divergence between the estimated variational goal posterior $\mathcal{Q}$ and a normal Gaussian distribution $\mathcal{N}(0,1)$: 
\begin{equation}
	\begin{aligned}
		\mathcal{L}_{goal} = & \alpha^{pred}_{goal}\left\|\boldsymbol{\overline{g}}-\boldsymbol{g}\right\|_1 \\
		& + \alpha^{kl}_{goal} \operatorname{KL}\left[\mathcal{Q}\left(\mathcal{Z}_{g}|\boldsymbol{f}_{ST}\right) \| \mathcal{N}(0,1)\right],
	\end{aligned}
\end{equation}  
where $\alpha^{pred}_{path}$ and $\alpha^{kl}_{path}$ balance the weight of $l_1$ error and KL divergence in $\mathcal{L}_{path}$.

\subsubsection{\textit{Path Decoder}: Plan Path towards Destination}
\revise{Based on the given text-scene joint condition $\boldsymbol{f}_{ST}$ and inferred goal $\boldsymbol{\overline{g}}$, path decoder $\mathit{\Theta}$ further plans a $N$-frame 3D body movement path $\boldsymbol{p}\in\mathbb{R}^{N\times3}$ towards the predicted goal $\boldsymbol{\overline{g}}$, where $\boldsymbol{p}_{n}$ denotes the 3D body center locations at $n$-th frame.} The predicted goal $\boldsymbol{\overline{g}}$ determines where to move to, while the joint conditions of textual instruction and 3D scene determine how to move. Therefore, given $\boldsymbol{\overline{g}}$ and $\boldsymbol{f}_{ST}$ as conditional inputs, path decoder $\mathit{\Theta}$ predicts a non-deterministic movement path $\boldsymbol{p}$ toward $\boldsymbol{\overline{g}}$ via exploring the uncertainty behind the posterior probability distribution $\mathcal{P}(\boldsymbol{p}|\overline{\boldsymbol{g}}, \boldsymbol{f}_{ST})$ as:
\begin{equation}
	\begin{aligned}
		 \boldsymbol{f}_p \sim & \mathcal{Q}(\mathcal{Z}_{p}|\boldsymbol{\overline{g}},\boldsymbol{f}_{ST}) \equiv \mathcal{N}(\boldsymbol{\mu}_{p}, \boldsymbol{\sigma}_{p}), \\
		\textit{where} \; \boldsymbol{\mu}_{p} & =\textit{\ttfamily{MLP}}_{p}^{1}(\boldsymbol{\overline{g}}||\boldsymbol{f}_{ST});\\ 
		 \boldsymbol{\sigma}_{p} & = \textit{\ttfamily{MLP}}_{p}^{2}(\boldsymbol{\overline{g}}||\boldsymbol{f}_{ST}). 
	\end{aligned}
\label{Eq.1}
\end{equation}
Similarly, as shown in Eq. \ref{Eq.1}, we first also concatenate $\boldsymbol{\overline{g}}$ and $\boldsymbol{f}_{ST}$ into an integrated representation $[\boldsymbol{\overline{g}}||\boldsymbol{f}_{ST}]$ and map it into two Gaussian distribution parameters $\boldsymbol{\mu}_{p}$ and $\boldsymbol{\sigma}_{p}$. Then, we parameterize a latent path feature space $\mathcal{Z}_{p}$ based on $\boldsymbol{\mu}_{p}$ and $\boldsymbol{\sigma}_{p}$. In this way, we estimate a variational path posterior $\mathcal{Q}$ by assuming a Gaussian information bottleneck. Finally, we sample a latent-based path feature $\boldsymbol{f}_{p}$ from $\mathcal{Z}_{p}$. \revise{After integrating $N$ sinusoidal positional embeddings into the sample latent $\boldsymbol{f}_{p}$, we send it into a Transformer-based path decoder $\mathit{\Theta}$ for decoding a $N$-frame 3D body motion path $\boldsymbol{\overline{p}}_{1:N}$ for a specified duration $N$ as:} 
\begin{equation}
	\boldsymbol{\overline{p}}_{1:N} = \mathit{\Theta}(\boldsymbol{f}_{p}).
\end{equation}
We also train $\mathit{\Theta}$ by minimizing two objectives: (1) the $l_1$ distance between the predicted 3D path $\boldsymbol{\overline{p}}_{1:N}$ and its ground-truth $\boldsymbol{p}_{1:N}$; (2) the Kullback-Leibler divergence between the estimated variational posterior $\mathcal{Q}$ and a normal Gaussian distribution $\mathcal{N}(0,1)$: 
\begin{equation}
\begin{aligned}
	\mathcal{L}_{path} = & \alpha^{pred}_{path}\left\|\boldsymbol{\overline{p}}_{1:N}-\boldsymbol{p}_{1:N}\right\|_1 \\
	& + \alpha^{kl}_{path} \operatorname{KL}\left[\mathcal{Q}\left(\mathcal{Z}_{p}|\overline{\boldsymbol{g}},\boldsymbol{f}_{T},\boldsymbol{f}_{S}\right) \| \mathcal{N}(0,1)\right].
\end{aligned}
\end{equation}

\subsubsection{\textit{Pose Decoder}: Synthesize Poses following Path}
Pose decoder $\mathit{\Psi}$ focuses on synthesizing realistic human poses following the inferred $N$-frame motion path $\boldsymbol{\overline{p}}$. Conditioned on $\boldsymbol{\overline{p}}$ and $\boldsymbol{f}_{ST}$, pose decoder explores diverse motion styles via learning the posterior probability distribution $\mathcal{P}(\boldsymbol{m}|\overline{\boldsymbol{p}},\boldsymbol{f}_{ST})$ as:  
\begin{equation}
	\begin{aligned}
		\boldsymbol{f}_m \sim & \mathcal{Q}(\mathcal{Z}_{m}|\boldsymbol{\overline{p}},\boldsymbol{f}_{ST}) \equiv \mathcal{N}(\boldsymbol{\mu}_{m}, \boldsymbol{\sigma}_{m}), \\
		\textit{where} \; \boldsymbol{\mu}_{m} & = \textit{\ttfamily{MLP}}_{m}^{1}(\boldsymbol{\overline{p}}||\boldsymbol{f}_{ST});\\ 
		\boldsymbol{\sigma}_{m} & = \textit{\ttfamily{MLP}}_{m}^{2}(\boldsymbol{\overline{p}}||\boldsymbol{f}_{ST}). 
	\end{aligned}
	\label{Eq.4}
\end{equation}
In Eq.\ref{Eq.4}, $\boldsymbol{\mu}_{m}$ and $\boldsymbol{\sigma}_{m}$ parameterize a latent pose feature space $\mathcal{Z}_{m}$ from which we sample a latent-based motion feature $\boldsymbol{f}_m$ for decoding as:
\begin{equation}
	\boldsymbol{\overline{m}}_{1:N} = \mathit{\Psi}(\boldsymbol{f}_{m}).
\end{equation}
Similarly, we train a Transformer-based pose decoder $\mathit{\Psi}$ also by minimizing $l_1$-based pose prediction error and KL-distance-based distribution regularization error:
\begin{equation}
	\begin{aligned}
		\mathcal{L}_{pose} = & \alpha^{pred}_{pose}\left\|\boldsymbol{\overline{m}}_{1:N}-\boldsymbol{m}_{1:N}\right\|_1 \\
		& + \alpha^{kl}_{pose} \operatorname{KL}\left[\mathcal{Q}\left(\mathcal{Z}_{m}|\boldsymbol{p},\boldsymbol{f}_{ST}\right) \| \mathcal{N}(0,1)\right].
	\end{aligned}
\end{equation} 

\subsection{Scene-Human Interaction Analyzer}
Given a synthesized human motion sample $\boldsymbol{\overline{m}}_{1:N}$ inside 3D scene $\boldsymbol{S}$, scene-aware motion analyzer recognizes its action category $\overline{\text{ACT}}$ and interaction object $\overline{\text{OBJ}}$ to scrutinize its semantic consistency between the given textual description $\boldsymbol{T}$. \revise{Specifically, we first deploy a pre-trained Mesh Transformer \cite{Zhu_2023_CVPR} as our body motion feature extractor. In \cite{Zhu_2023_CVPR}, it pre-trains Mesh Transformer via two self-supervised tasks, namely masked vertex modeling and future frame prediction. Therefore, as a powerful motion feature extractor, this pre-trained Mesh Transformer embeds $\boldsymbol{\overline{m}}_{1:N}$ into its $M$-token body embedding sequence $\boldsymbol{f}_{B}=[\boldsymbol{f}_{B}^{1}, \cdots, \boldsymbol{f}_{B}^{M}]$.} Then, we develop a series of self-attention and cross-attention layers between the given extracted body motion feature $\boldsymbol{f}_{B}$ and scene feature $\boldsymbol{f}_{S}$ to infer $\overline{\text{ACT}}$ and $\overline{\text{OBJ}}$ from them.   

For simplicity, we take the operation in $l$-th self-attention and cross-attention layers as an example. We stack $L$ such layer to form a scene-human interaction analyzer. At $l$-th layer, self-attention layer takes $\boldsymbol{f}_{B}^{l}$ as input and updates it into $\boldsymbol{f}_{B}^{l+1}$ as: $\boldsymbol{f}_{B}^{l+1}=\textit{\ttfamily{SelfAtt}}(\boldsymbol{f}_{B}^{l})$. Concurrently, cross-attention layer characterizes the body-scene interaction dependencies between $\boldsymbol{f}_{B}^{l}$ and $\boldsymbol{f}_{S}^{l}$ as: $\boldsymbol{f}_{S}^{l+1}=\textit{\ttfamily{CrossAtt}}(\boldsymbol{f}_{B}^{l}, \boldsymbol{f}_{S}^{l})$. After iterating these operations $L$ times, we then send $\boldsymbol{f}_{B}^{L}$ into a action classifier to infer the probability distribution $\overline{P}_{ACT}$ over all action categories as: 
\begin{equation}
\overline{P}_{ACT} = \textit{\ttfamily{SoftMax}}\left(\textit{\ttfamily{MLP}}\left(\boldsymbol{f}_{B}^{L}\right)\right).  
\end{equation}  
We represent the ground-truth of $\overline{P}_{ACT}$ as a one-hot vector $P_{ACT}$, whose value on the index of ground-truth ${\text{ACT}}$ is one and the others are all zeros.

Similarly, given the human-scene interaction embedding $\boldsymbol{f}_{S}^{L}$, we infer the indoor object that $\boldsymbol{\overline{m}}_{1:N}$ interacts with inside the 3D scene $\boldsymbol{S}$ as:
\begin{equation}
	\overline{P}_{OBJ} = \textit{\ttfamily{SoftMax}}\left(\textit{\ttfamily{MLP}}\left(\boldsymbol{f}_{S}^{L}\right)\right),  
\end{equation}
where $\overline{P}_{OBJ}$ denotes the inferred probability distribution over all indoor objects and its ground-truth is one-hot vector $P_{OBJ}$. Similar to the definition of $P_{ACT}$, $P_{OBJ}$ defines the value on the index of ground-truth $\text{OBJ}$ as one and others as zeros.  
 
We train the human-scene interaction analyzer with minimizing the cross-entropy-based classification loss between inferred action probability distribution $\overline{P}_{OBJ}$, interaction object probability distribution $\overline{P}_{OBJ}$, and their ground-truths as:
\begin{equation}	
\mathcal{L}_{rec} = \textit{\ttfamily{CE}}(\overline{P}_{ACT}, P_{ACT}) + \textit{\ttfamily{CE}}(\overline{P}_{OBJ}, P_{OBJ}),
\end{equation} 
where \textit{\ttfamily{CE}}($\cdot$) calculates the cross-entropy between two probability distributions.

Finally, we integrate all training losses and optimize all components ($i.e.$, motion generator and analyzer) end-to-end as:
\begin{equation}	
	\mathcal{L} = \alpha_{goal}\mathcal{L}_{goal} + \alpha_{path}\mathcal{L}_{path} + \alpha_{pose}\mathcal{L}_{pose} + \alpha_{rec}\mathcal{L}_{rec},
\end{equation} 
where $\alpha$ denotes the loss weight of each term.

\subsection{Discussion}
\label{sec:discuss}

\mr{In this section, we give more in-depth analyses of our proposed CESA scheme. Specifically, compared with the original GAN-based generation strategy \cite{goodfellow2020generative}, our CESA decreases its model collapse risk from three aspects: (1) \textit{three-stage inference strategy for diverse generation}. The goal-path-pose three-stage inference scheme encourages the text-to-motion generator to improve the intra-class diversity of synthesized samples; (2) \textit{multi-semantic recognition strategy for versatile discrimination}. Different from the true/false two-category recognition between real and synthesized samples, CESA enforces a stronger fine-grained semantic discrimination on synthesized samples by recognizing their action categories and interaction objects; (3) \textit{multi-object loss function for strong constraint}. CESA is optimized with a multi-object loss function that stabilizes the training by minimizing goal-path-pose inference and action-object recognition errors. Decoupling these three proposals, CESA outperforms original GAN-based synthesis methods with better training stability and fewer model collapse risks.}
    
\begin{table*}[t]
	\centering
		\caption{Quantitative comparisons of scene-aware text-to-motion generation on HUMANISE. The best results are marked in bold.}
	\scalebox{1}{
		\begin{tabular}{lccccccc}
			\toprule
			Methods & \; FID $\downarrow$ \quad & \; \mr{DIV} $\uparrow$ \quad & \quad MRS $\uparrow$ \quad & \quad TMCS $\uparrow$ \quad &\; ACC $\uparrow$ \quad& Non-collision$\uparrow$ & Contact$\uparrow$ \\ \hline
			LTMI  \cite{wang2020synthesizing} & $3.237$    	& $6.51$    & $2.52$    & $3.02$     & $0.911$  &  $98.22$                 & $ 80.15$          \\
			T2M-Scene \cite{wang2022humanise}	& $3.125$    	&  $6.72$   & $2.57$    &  $3.28$    & $0.924$  &   $98.21$                &   $80.13$            \\
			MLD \cite{chen2023executing} 		& $3.012$    	& $5.31$    & $2.60$    & $3.61$     & $0.911$  &  $98.21$                 & $ 80.14$  \\
			DIMOS \cite{ZhaoICCV2023}	& $2.914$    	& $10.51$    & $6.45$    & $3.59$     & $0.917$  &  $98.64$                 & $ 80.37$          \\
			COINS \cite{zhao2022compositional}	& $2.739$    	& $5.11$    & $3.60$    & $3.71$     & $0.937$  &  $98.77$       & $ 80.40$          \\
			GUESS \cite{gao2024guess}    & $2.691$    	& $6.42$    & $3.64$    & $3.73$     & $0.939$  &  $98.69$       & $ 80.64$          \\ 
		    T2M-GPT \cite{zhang2023t2m} & $2.831$    	& $6.73$    & $3.37$    &  $3.66$    & $0.933$  & $98.56$ &  $80.20$          \\ 
			GenZI \cite{li2024genzi} & $2.503$    	& $6.82$    & $3.84$    &  $4.02$    & $0.948$  & $99.01$ &  $81.89$ \\ 
			Act2HSI \cite{jiang2024scaling} & $2.437$    	& $7.31$    & $3.98$    &  $4.11$    & $0.959$  & $99.03$ &  $82.09$ \\ 
			AffordMotion \cite{wang2024move} & $2.416$    	& $7.56$    & $4.04$    &  $4.16$    & $0.961$  & $99.05$ &  $82.18$          \\  
			\hline
			CESA	(\textit{synthesis-only})	& \textcolor{black}{$2.663$}    	& \textcolor{black}{$7.81$}    &  \textcolor{black}{$3.69$}   &  \textcolor{black}{$3.77$}    &\textcolor{black}{$0.937$}   & \textcolor{black}{$98.92$}                & \textcolor{black}{$81.61$}                                 \\
			CESA	(\textit{synthesis \& analysis})  & $\boldsymbol{2.005}$ & $\boldsymbol{7.89}$    & $\boldsymbol{4.16}$   & $\boldsymbol{4.39}$    & $\boldsymbol{0.998}$  & $\boldsymbol{99.14}$                &$\boldsymbol{83.16}$                                \\ \bottomrule
		\end{tabular}
	}
\vspace{0.1in}
	\label{table1}
\end{table*}

\begin{table*}[t]
	\centering
	\caption{Quantitative comparisons of scene-aware text-to-motion generation on TRUMANS. The best results are marked in bold.}
	\scalebox{1}{
		\begin{tabular}{lccccccc}
			\toprule
			Methods & \; FID $\downarrow$ \quad & \; \mr{DIV} $\uparrow$ \quad & \quad MRS $\uparrow$ \quad & \quad TMCS $\uparrow$ \quad &\; ACC $\uparrow$ \quad& Non-collision$\uparrow$ & Contact$\uparrow$ \\ \hline
			T2M-Scene \cite{wang2022humanise}	     & $1.932$    	&  $5.69$   & $3.42$    &  $3.21$    & $0.803$  &   $97.31$       & $79.35$            \\
			COINS \cite{zhao2022compositional}	     & $2.033$    	&  $6.14$    & $3.36$    & $3.42$     & $0.762$  &  $97.02$       & $77.96$          \\     
			GenZI \cite{li2024genzi}                 & $1.878$    	& $ 6.33$    & $3.81$    &  $3.79$    & $0.825$  & $97.69$ &  $80.45$ \\ 
			AffordMotion \cite{wang2024move}         & $1.599$    	& $6.42$    & $3.92$    &  $4.02$    & $0.842$  & $98.30$ &  $81.54$          \\  
			Act2HSI \cite{jiang2024scaling}          & $1.025$    	& $6.30$    & $4.09$    &  $4.21$    & $0.861$  & $\boldsymbol{99.11}$ &  $83.33$ \\ 
			\hline
			CESA	(\textit{synthesis-only})	& $1.477$    	& $6.73$    &  $4.05$   &  $4.19$    &\textcolor{black}{$0.853$}   & \textcolor{black}{$98.65$}                & \textcolor{black}{$81.96$}                                 \\
			CESA	(\textit{synthesis \& analysis})  & $\boldsymbol{1.002}$ & $\boldsymbol{6.80}$    & $\boldsymbol{4.36}$   & $\boldsymbol{4.45}$    & $\boldsymbol{0.913}$  & $99.09$                &$\boldsymbol{83.62}$                                \\ \bottomrule
		\end{tabular}
	}
	\vspace{0.1in}
	\label{table_TRUMANS}
\end{table*}


\section{Experiments and Analysis}
\subsection{Dataset} 

$\quad$ \textbf{HUMANISE} \cite{wang2022humanise} is currently largest text-annotated indoor human motion dataset that captures 19.6k 3D human motion clips inside 643 3D scenes and annotates each motion-scene pair with a textual description. The collected actions include sitting, standing up, walking, \textit{etc}., -- a set of basic daily activities within indoor scenes. Besides, the scanned 3D scenes include living rooms, bedrooms, kitchens, balconies, \textit{etc.}, enabling humans to interact with diverse indoor objects. These human-scene interaction samples encode meaningful and versatile object/scene affordances, which introduce both potentials and challenges to the model building. Furthermore, each motion-scene pair is annotated by an English-based sentence that describes its action type and the objects/locations being interacted with. 

\textbf{TRUMANS} \cite{jiang2024scaling} encompasses over 15 hours of diverse human interactions across 100 indoor scenes, comprising a total of 1.6 million frames. The human-scene interactions in TRUMANS include 20 different types of common objects, ensuring a minimum of 5 distinct instances per type. These scenes span a variety of settings, such as dining rooms, living rooms, bedrooms, and kitchens, among others. TRUMANS also annotates each human-scene interaction pair with a frame-wise action label.

\textbf{PROX-S} \cite{zhao2022compositional} annotates each 3D body-scene interaction sequence of PROX \cite{hassan2019resolving} with semantic language descriptions. The PROX-S dataset contains (1) 3D instance segmentation of all 12 PROX scenes; (2) 3D human body reconstructions within the scenes; and (3) per-frame textual annotation in the form of action-object pairs. As a large-scale text-annotated body-scene interaction dataset, PROX-S collects around 32K frames of 3D human-scene interactions from 43 sequences recorded in 12 indoor scenes.

\textbf{Sketchfab} \cite{li2024genzi} gathers 8 large-scale 3D scenes encompassing a variety of indoor and outdoor environments with diverse geometric structures, including a realistic Venice city, a gym, and a cartoon-style food truck. It collects 4-5 text prompts per scene describing human interactions with the scene for specified approximate point locations, resulting in 38 actions for evaluation.


\subsection{Implementation Details}
In the language-condition human motion synthesis inside 3D scenes, GoalDecoder $\mathit{\Phi}$ is a 2-layer Transformer with 2 heads. PathDecoder $\mathit{\Theta}$ is a 4-layer Transformer with 4 heads. PoseDecoder $\mathit{\Psi}$ is a 4-layer Transformer with 4 heads. In the human motion analysis, we stack 4 self-attention layers and 4 cross-attention layers (\textit{i.e.}, $L$=4). The channel dimension of feature embeddings (\textit{i.e.}, $\boldsymbol{f}_{T}$, $\boldsymbol{f}_{S}$, $\boldsymbol{f}_{g}$, $\boldsymbol{f}_{p}$, $\boldsymbol{f}_{m}$, and $\boldsymbol{f}_{B}$) are both 512. The channel dimensions of Gaussian parameters $\boldsymbol{\mu}$ and $\boldsymbol{\sigma}$ are both 32. \revise{In the training stage, we fix the parameters of official pre-trained Point transformer \cite{zhao2021point}, BERT \cite{kenton2019bert} and Mesh Transformer \cite{Zhu_2023_CVPR}. All learnable components are trained end-to-end. There parameters are optimized by Adam \cite{kingma2014adam} with fixed learning rate 0.001, training epoch 150, and batch size 32. The hyper-parameters in training loss are set as: $\alpha_{goal}=1, \alpha_{path}=1, \alpha_{pose}=1, \alpha_{rec}=10, \alpha^{pred}=1, \alpha^{kl}=0.1$. Finally, we implement the training and inference stages of CESA with PyTorch 1.7 on one RTX-3090Ti GPU.}

\subsection{Compared Methods}
\mr{The baseline methods we compared can be divided into four model groups: (1) \textit{scene-aware text-driven motion sequence synthesis models.} \textit{T2M-Scene} \cite{wang2022humanise}, \textit{AffordMotion} \cite{wang2024move}, and \textit{Act2HSI} \cite{jiang2024scaling} are most related works, to our knowledge. They develop three strong baselines on text-to-motion synthesis in 3D scenes; (2) \textit{scene-aware text-driven body pose synthesis models.} While \textit{COINS} \cite{zhao2022compositional} and \textit{GenZI} \cite{li2024genzi} are another two attempts at scene-aware text-to-body synthesis, they generate a single 3D body pose rather than a 3D body motion sequence. We thus tune them into a sequence-level generation; (3) \textit{scene-aware waypoint-driven motion sequence synthesis models.} We also spend extensive efforts to develop \textit{DIMOS} \cite{ZhaoICCV2023} and  \textit{LTMI} \cite{wang2020synthesizing} from waypoint-driven body-scene interaction synthesis systems into a text-driven ones; (4) \textit{text-driven motion sequence synthesis models.} \textit{T2M-GPT} \cite{zhang2023t2m}, MLD \cite{chen2023executing}, and GUESS \cite{gao2024guess} are state-of-the-art text-to-motion synthesis methods. We thus introduce 3D scene contexts into their condition inputs. \textit{We tune all these baseline methods based on their official codes. Please refer to the appendix for more implementation details.}} 

\begin{table*}[t]
	\centering
	\caption{Quantitative comparisons of text-to-motion generation on Sketchfab and HumanML3D datasets. The best results are marked in bold.}
	\scalebox{1}{
		\begin{tabular}{lccccccccc}
			\toprule
			\multirow{2}{*}{Methods} & \multicolumn{4}{c}{PROX-S} &  & \multicolumn{4}{c}{Sketchfab} \\ \cline{2-5} \cline{7-10} 
			& \; FID $\downarrow$  \;& \mr{DIV} $\uparrow$  & MRS $\uparrow$   &  TMCS $\uparrow$  & \qquad &  \; FID $\downarrow$ \; &  \mr{DIV} $\uparrow$  & MRS $\uparrow$ & TMCS $\uparrow$ \\ \hline
			T2M-Scene \cite{wang2022humanise} &  $2.004$ & $7.44$                & $3.59$     & $3.68$      &  &  $0.871$    & $5.78$   &  $3.48$    & $3.87$     \\
			COINS \cite{zhao2022compositional}&  $1.721$ & $7.31$                & $4.16$     & $4.27$      &  &  $0.791$    & $6.20$   &  $3.98$    & $4.07$ \\
			GenZI \cite{li2024genzi}          &  $1.511$ & $7.20$                &  $4.21$    & $4.29$      &  &  $0.644$    & $6.14$   & $6.11$     & $4.10$ \\ 
			AffordMotion \cite{wang2024move}  &  $1.401$ &  $7.59$               &  $4.31$    & $4.33$      &  &  $0.596$    & $6.29$   & $4.22$     & $4.19$ \\ 
			Act2HSI \cite{jiang2024scaling}   &  $1.451$ &  $7.28$               &  $4.26$    & $4.22$      &  &  $0.622$    & $6.33$   & $4.09$     & $4.11$ \\
			\hline
			CESA (\textit{synthesis-only})  	  &  $1.694$ & $7.79$  & $4.02$      & $4.17$      &  &  $0.613$   & $6.51$    &   $4.02$   & $4.07$ \\
			CESA (\textit{synthesis \& analysis}) \qquad \qquad & $\boldsymbol{1.374}$ &  $\boldsymbol{7.86}$     & $\boldsymbol{4.53}$ & $\boldsymbol{4.45}$      &  & \textcolor{black}{$\boldsymbol{0.587}$}  &  $\boldsymbol{6.60}$   & \textcolor{black}{$\boldsymbol{4.27}$}     & $\boldsymbol{4.31}$  \\ \bottomrule
		\end{tabular} 
	}
	\vspace{0.1in}
	\label{table3_prox}
\end{table*}

\begin{figure*}[t]
	\centering
	\includegraphics[width=1\textwidth]{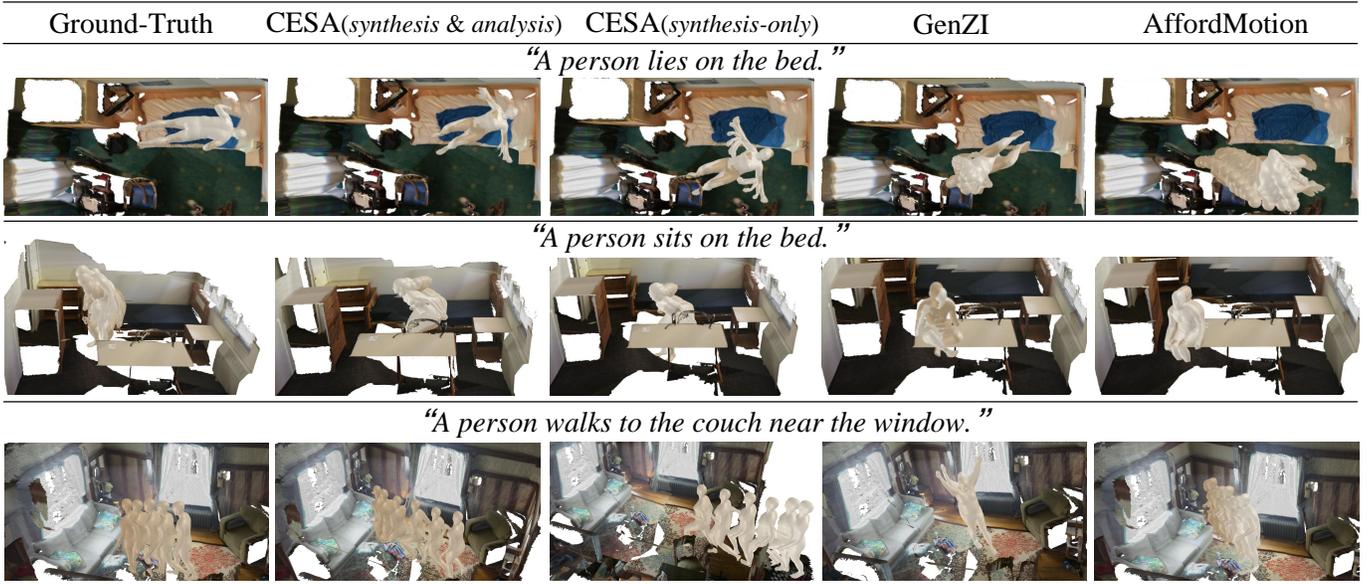}	
	\caption{Qualitative Comparison. We visualize four human-scene interaction samples synthesized from each text-to-motion generation method. Dotted boxes indicate the imperfections reflected in poor motion naturalness, unrealistic human-scene interactions, and mistaken interaction objects. The color of the pose deepens over time.}
	\label{VIS_1}
\end{figure*}

\subsection{Evaluation Metrics}
\textbf{Generation Metrics}. \revise{We first adopt five quantitative metrics that are widely used in \cite{guo2022generating, guo2022tm2t, zhang2023t2m} to comprehensively evaluate synthesized body motion samples for their realism, diversity, text-motion consistency, and motion-scene compatibility. In the following, we introduce the definitions of these metrics.}
\begin{itemize}
	\item \textit{Frechet Inception Distance} (FID) reflects the realism of synthesized motions. \revise{It evaluates the latent-based feature distribution distance between the generated and real motions as: $\text{FID} = \left\|\mu_{gt}-\mu_{gen}\right\|^2-\operatorname{Tr}\left(\Sigma_{gt}+\Sigma_{gen}-2\left(\Sigma_{gt} \Sigma_{gen}\right)^{\frac{1}{2}}\right)$, where $f_{gt}$ and $f_{gen}$ are ground-truth and generated motion features, respectively. They are extracted with pre-trained networks in \cite{guo2022generating}.}
	\mr{\item \textit{Diversity} (DIV) reflects the diversity within the set of synthesized motion samples. From a set of all generated motions from various descriptions, two sub-sets of the same size $S$ are randomly sampled. Their respective sets of extracted motion feature vectors are ${\mathbf{v}_1,\cdots,\mathbf{v}_S}$ and ${\mathbf{v}_1^{\prime},\cdots,\mathbf{v}_S^{\prime}}$. Following \cite{guo2022generating,guo2022tm2t,jiang2023motiongpt} DIV is defined as: $\text{DIV}=\frac{1}{S}\sum_{i=1}^{S} \left\|\mathbf{v}_i-\mathbf{v}_i^{\prime}\right\|$.}
	\item \textit{Non-Collision Score} and \textit{Contact Score} are in-environment evaluation metrics and reflect motion-scene compatibility \cite{zhang2020generating, wang2021synthesizing}.    
	\item \textit{Accuracy} (ACC) is computed as the average action recognition performance with generated motions, reflecting their fidelities with action types specified in textual conditions.
\end{itemize}

\mr{\textbf{Perceptual Study}. We also perform user perceptual studies to intuitively evaluate the generation results in terms of their overall quality and action-semantic accuracy performances. Similar to \cite{wang2024move, wang2022humanise}, we generate 20 human motion samples conditioned on 20 text-scene scenarios for each model. Given each language description and synthesized human motion pair, a participant is asked to respectively score from 1 to 5 for (i) the overall motion realism quality, and (ii) the semantic consistency with a given text command. A higher rated score indicates that the generated result is more realistic and plausible. Then, we respectively average the scores of all 40 participants to obtain the Motion Realism Score (MRS) and Text-Motion Consistency Score (TMCS) evaluation performances of each model.}

\begin{figure*}[t]
	\centering
	\includegraphics[width=1\textwidth]{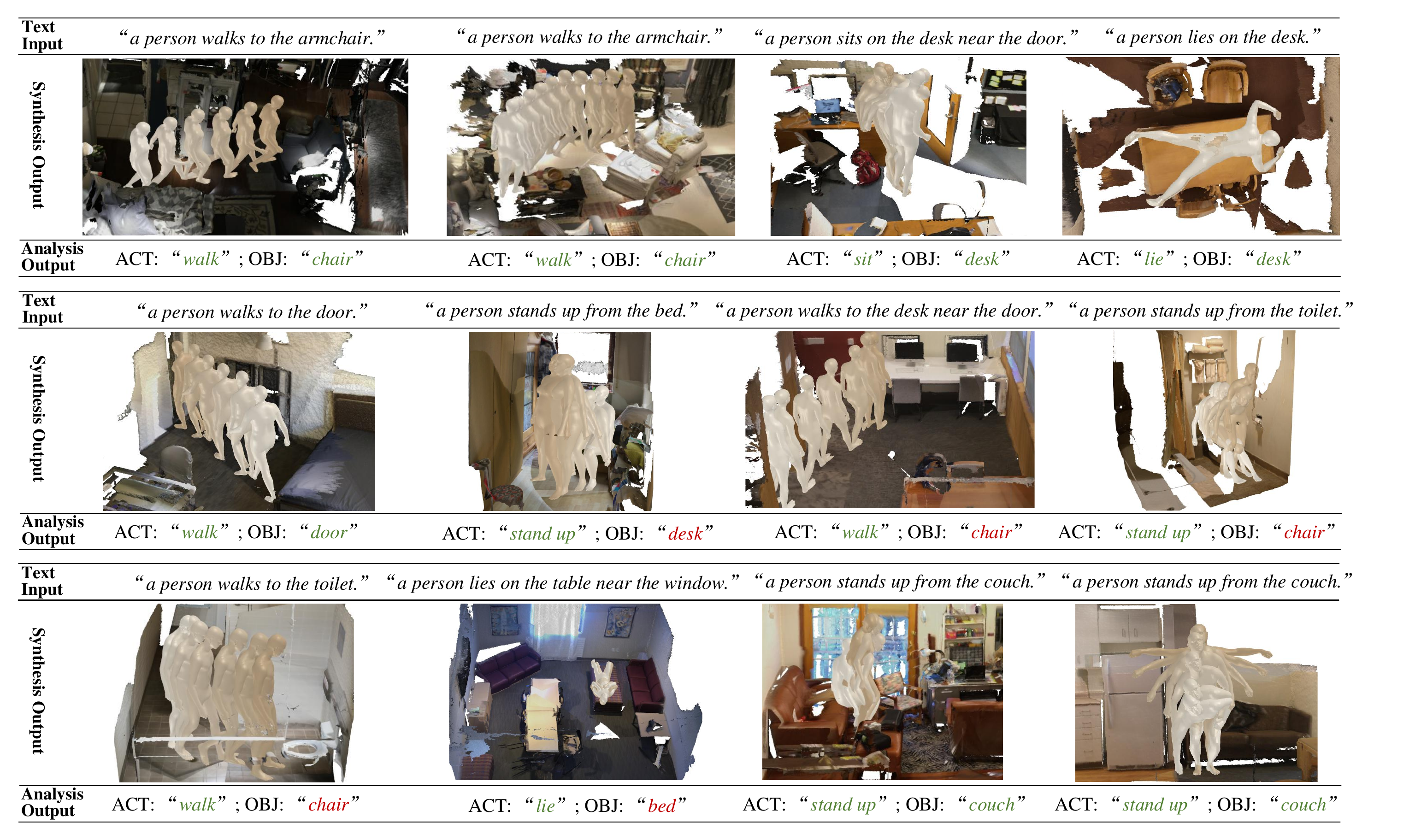}
	\caption{Motion synthesis and analysis results conditioned on the given text inputs. We indicate the correct and incorrect inferred results of ACT and OBJ with \textcolor{green}{green} and \textcolor{red}{red}, respectively.}
	\label{VIS_more}
\end{figure*}
\subsection{Performance Comparison}
\subsubsection{Quantitative Comparison}
In this section, we analyze the performance of our and previous methods with quantitative comparisons. Specifically, as shown in Tab. \ref{table1}, our method significantly outperforms state-of-the-art models on HUMANISE dataset by large margins: 35\% on FID, 18\% on MRS, and 19\% on TMCS, $etc$. Besides, as shown in Tab. \ref{table_TRUMANS} and Tab. \ref{table3_prox}, CESA also obtains better performances on TRUMANS, PROX-S, and Sketchfab datasets. All these quantitative performance gains on four datasets verify that our synthesized human motions have better realism, diversity, and text-motion consistency. Furthermore, we also report the performance of the synthesis-only module to investigate the effects of the human motion analysis branch on text-to-motion generation in 3D scenes. Quantitative performance gaps between these two different model configurations verify that integrating motion synthesis and analysis into a co-evolving pipeline significantly improves motion realism and text-motion consistency.    

\subsubsection{Qualitative Comparison}
In this section, we evaluate the performance of different scene-aware text-to-motion generation methods with qualitative comparisons \footnote{Please refer to the demo video for more visualizations.}. Quantitative results (Tab. \ref{table1}) indicate that integrating motion analysis with synthesis will enforce quality scrutiny on generated human motions, significantly improving realism and text-motion consistency. Results shown in Fig. \ref{VIS_1} also verify that the human-scene interaction analysis branch alleviates the imperfection of motions generated by the synthesis-only model, such as poor motion naturalness, unrealistic human-scene interactions, and mistaken interaction objects. Visually, given the same textual command and 3D scene inputs, the human-scene interactions generated by our CESA are more realistic and highly consistent with their textual semantics, outperforming other methods on fidelity by a large margin.   
\begin{figure*}[t]
	\centering
	\includegraphics[width=1\textwidth]{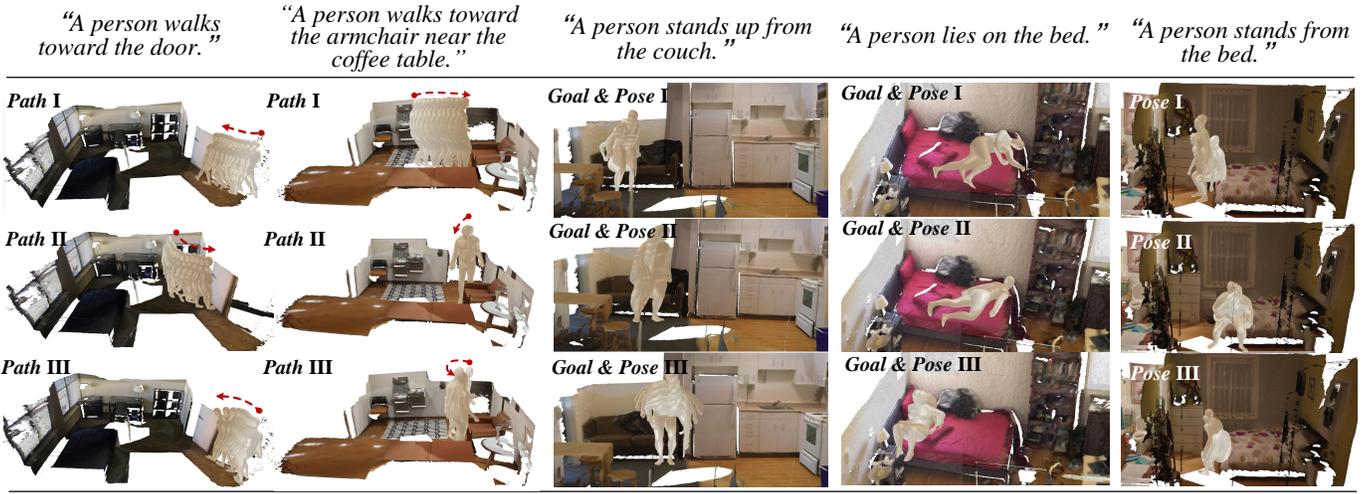}
	\caption{Diverse Text-to-Motion Generation. These visualizations of synthesized human-scene interaction samples reflect their diversities of movement paths and pose styles. Red dotted lines indicate their movement trajectories. The color of the pose deepens over time.}
	\label{VIS_2}
\end{figure*} 

\subsection{Sample Visualization}
In this section, we visualize several different human motion results synthesized from the motion generator and report their recognition results inferred from the motion analyzer. As shown in Fig. \ref{VIS_more}, CESA synthesizes realistic indoor human motion samples that are both compatible with their given scene contexts and consistent with their given textual descriptions. Besides, we further respectively report the ACT and OBJ recognition results inferred from each synthesized human motion sample and its given scene context. Notably, compared with ACT recognition, OBJ recognition is a more challenging joint inference based on the 3D motion and 3D scene and also should be robust to diverse object shapes, colors, and textures. Therefore, as shown in Fig. \ref{VIS_more}, similar shapes among some different object semantics (\textit{e.g.}, \textit{chair} and \textit{toilet}, \textit{bed} and \textit{desk}) would detract the robustness of OBJ recognition. 

\subsection{Motion Diversity}
In this section, we visualize multiple human-scene interaction samples synthesized from a same text-scene condition pair to intuitively evaluate our performances on motion diversity. Specifically, we take five different text-scene pairs as input examples and visualize three human motion samples generated from each condition input. As shown in Fig. \ref{VIS_2}, we can see that our method can generate several plausible motions conditioned on the same given textual description and scene context, performing diverse human motion generation. For example, given the same textual command ``walk to the door", we synthesized diverse walking samples with different movement paths but towards the same movement destination. Similarly, we also synthesized various ``standing up" actions with different motion styles. These visualizations indicate that CESA explores the freedom behind given linguistic descriptions and significantly improves generation diversity on different motion attributes ($e.g.$, movement path and pose style).  

 \begin{figure}[t]
	\centering
	\includegraphics[width=0.4\textwidth]{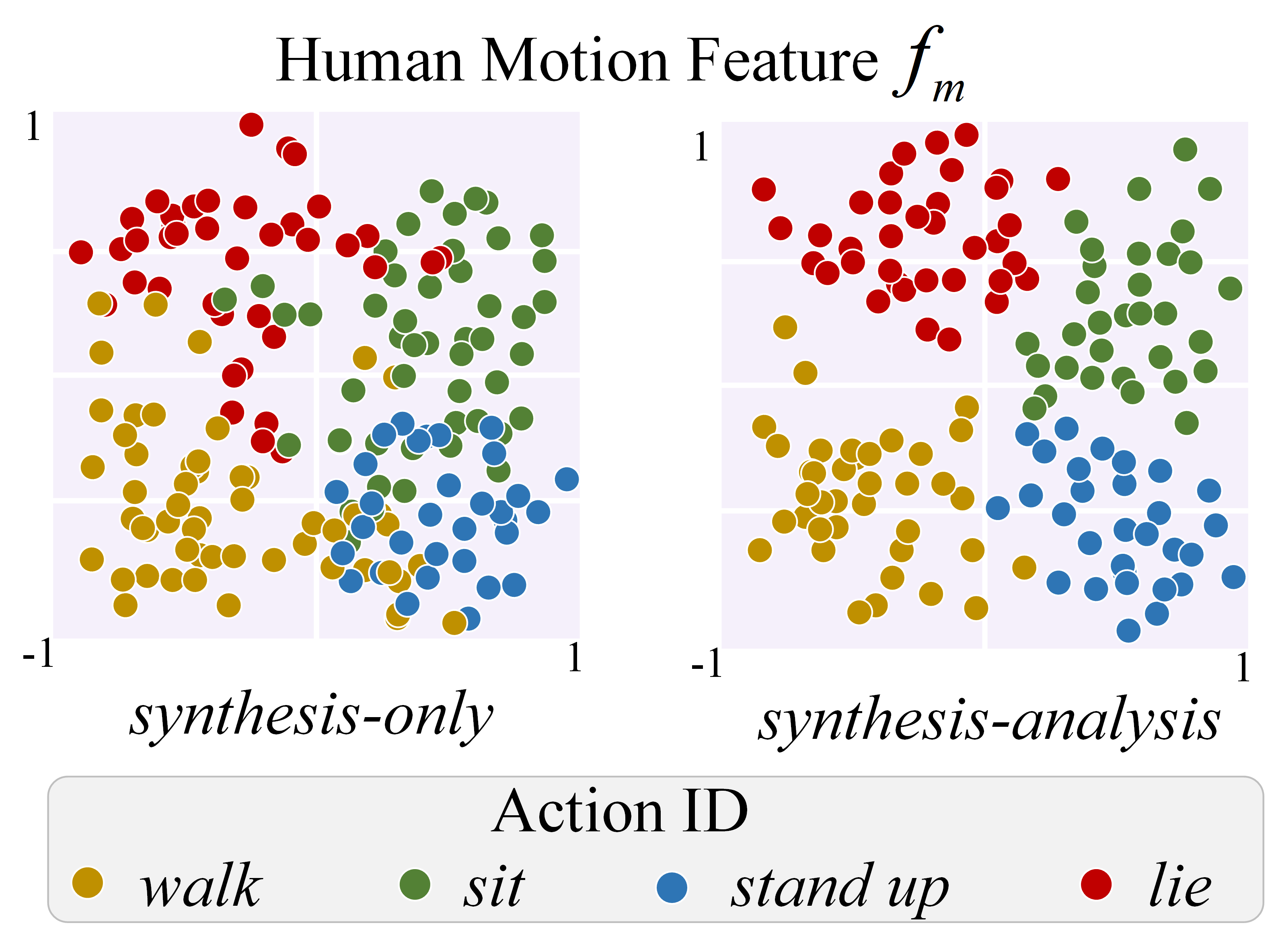}	
	\caption{t-SNE visualization of human motion features $\boldsymbol{f}_m$ generated from synthesis-only and synthesis-analysis setups.}
	\label{VIS_3}
\end{figure}

\begin{table}[t]
\centering
\caption{Performance comparisons between different ablative configurations of the motion analysis branch.}
\scalebox{0.7}{
\begin{tabular}{c|c|c|c|c|c}
\toprule
\multicolumn{1}{c|}{\multirow{2}{*}{Motion Synthesis}} & \multicolumn{2}{c|}{Motion Analysis}                   & \multicolumn{1}{c|}{\multirow{2}{*}{FID $\downarrow$}} & \multicolumn{1}{c|}{\multirow{2}{*}{MRS $\uparrow$}} & \multirow{2}{*}{TMCS $\uparrow$} \\ \cline{2-3}
\multicolumn{1}{c|}{}                                  & ACT Analysis & \multicolumn{1}{c|}{OBJ Analysis} & \multicolumn{1}{c|}{}                     & \multicolumn{1}{c|}{}                     &                       \\ \hline
$\checkmark$ &    $\times$   &   $\times$   & $2.663$ & $3.69$ & $3.77$ \\
$\checkmark$&    $\times$&    $\checkmark$ & $2.531$  & $3.82$ & $4.11$      \\ 
$\checkmark$&   $\checkmark$& $\times$  &  $2.250$    & $3.95$ & $3.89$  \\ 
$\checkmark$ &  $\checkmark$& $\checkmark$  & $2.005$ & $4.16$ & $4.39$ \\ \bottomrule              
\end{tabular}
}
\vspace{0.1in}
\label{tab.act&obj}
\end{table}

\subsection{Ablation Study}
In this section, we analyze the individual components and investigate their effects on the final system.

\subsubsection{Effect of Motion Analysis on Synthesis}
As verified in former sections (Tab. \ref{table1}, and Fig. \ref{VIS_1}), integrating motion synthesis and analysis into CESA brings quantitative and qualitative performance gains to text-to-motion generation in 3D scenes. In this section, we further investigate the effect of motion analysis on synthesis with more experimental results. As shown in Fig. \ref{VIS_3}, we first report the t-SNE visualization of synthesized latent-based human motion features $\boldsymbol{f}_m$. Figure \ref{VIS_3} (left) shows that without motion analysis, the human motions generated from the synthesis-only model are over-divergent. In this case, the synthesis-only setup requires the quality scrutiny of its generated human motions to improve text-motion semantic consistency and motion realism. Then, we further investigate the individual effects of ACT and OBJ analyses on motion synthesis. As shown in Tab. \ref{tab.act&obj}, performing ACT analysis introduces more performance gains on FID and MRS. In contrast, introducing OBJ analysis is more beneficial to improve TMCS performance. All these quantitative and qualitative experiments both verify that human motion analysis significantly benefits scene-aware text-to-motion generation by improving its motion realism and text-motion consistency.

 \begin{figure}[t]
	\centering
	\includegraphics[width=0.45\textwidth]{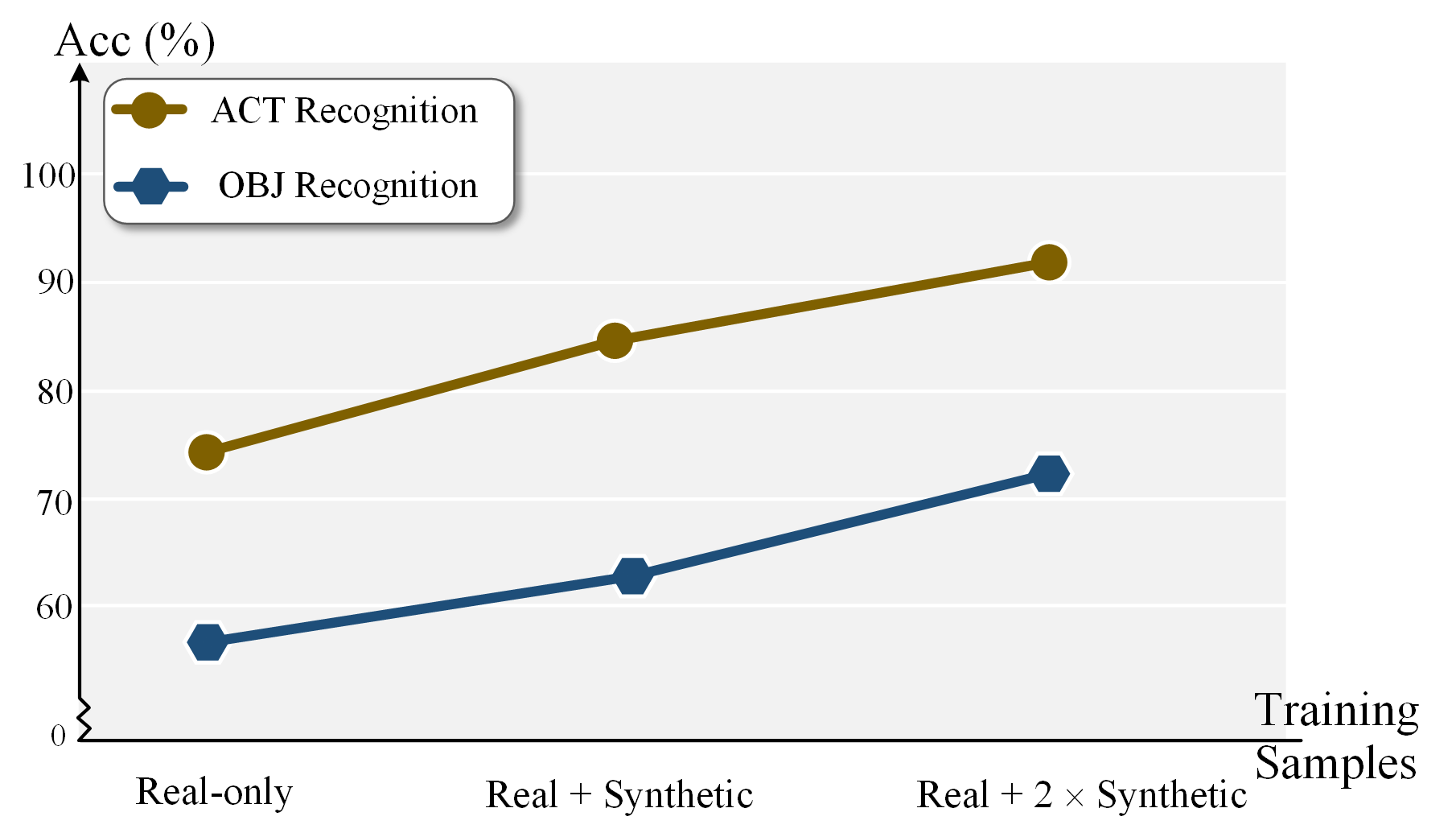}	
	\caption{The action category and interaction object recognition performance comparisons between different training setups. Synthetic human motion samples improve action and object recognition performances via enriching intra-class diversity.}
	\label{VIS_4}
\end{figure}

 \begin{figure*}[t]
	\centering
	\includegraphics[width=0.97\textwidth]{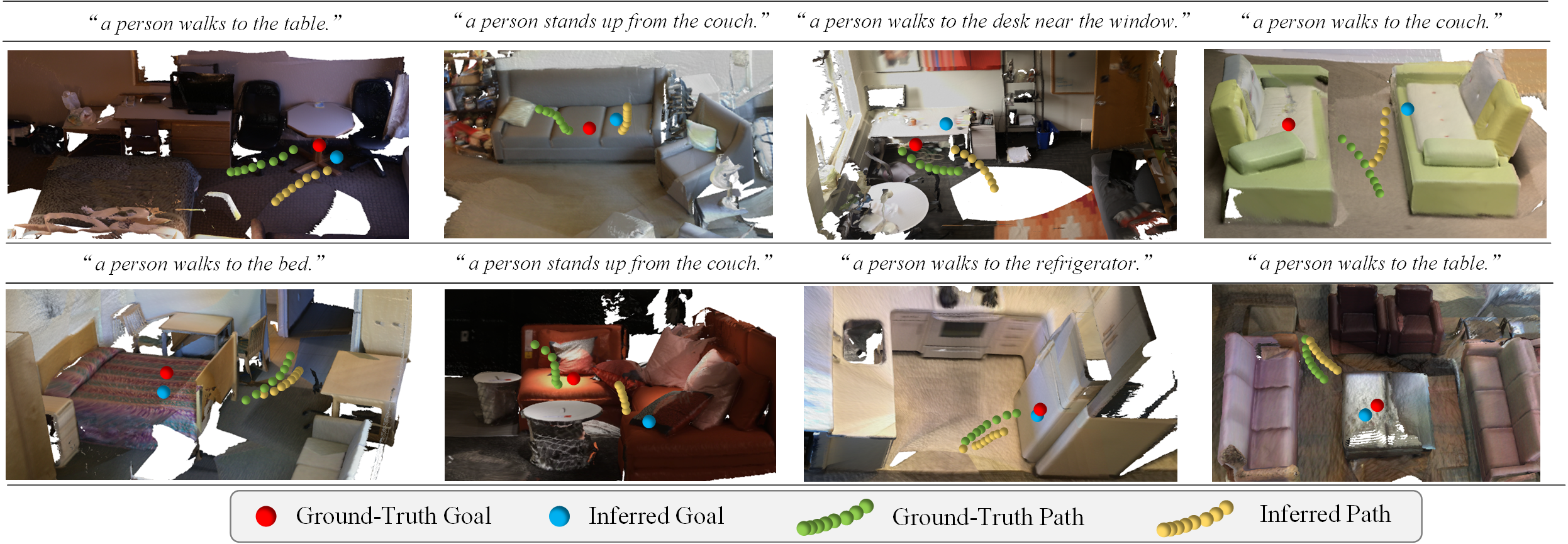}	
	\caption{Visualization comparison between inferred paths and goals with their ground-truths. }
	\label{VIS_5}
\end{figure*}

\subsubsection{Effect of Motion Synthesis on Analysis}
In this section, we investigate the effect of human motion synthesis on analysis within CESA. Specifically, in this ablation component study, we respectively evaluate the recognition accuracy performances of the human-scene interaction analysis model trained with three different sample sets. As shown in Fig. \ref{VIS_4}, as a baseline setup, we first use real 3D body motion samples of the HUMANISE training set to train a human-scene interaction analysis model. Then, we double the training sample size by introducing synthetic 3D body motion samples generated from the textual instructions of the HUMANISE training set. Besides, we repeat text-to-motion synthesis twice and thus triple the training sample size. Figure \ref{VIS_4} verifies that synthesized motion samples enrich the intra-class diversity of human-scene interactions and thus improve the recognition performance of action categories and interaction objects, significantly benefiting robust human-scene interaction analysis. Furthermore, Figure \ref{VIS_4} also indicates that more synthetic human-scene interaction samples tend to bring better recognition performance.

\begin{table}[t]
	\centering
	\caption{Ablative studies of GoalDecoder and PathDecoder on pose synthesis. Because of non-deterministic path inference, we repeat the path evaluation 20 times and report the average with 95\% confidence interval. Goal/Path errors are average 3D distances in meters.}
	\resizebox{0.9\columnwidth}{!}{
		\begin{tabular}{c|c|c|c|c}
			\toprule
			\multicolumn{2}{c|}{CESA Configuration} & \multicolumn{3}{c}{Pose Synthesis}  \\ \hline
			GoalDecoder          & PathDecoder          & FID $\downarrow$ & Goal Error $\downarrow$ & Path Error $\downarrow$ \\ \hline
			$\times$ &  $\times$                    & $2.512$ &  $1.47$    & $2.31$              \\ 
			$\times$ &  $\checkmark$                & $2.198$ &  $1.21$    & $1.84$               \\
			$\checkmark$ &  $\times$                & $2.451$ &  $0.89$    & $1.93$               \\
			$\checkmark$ &  $\checkmark$            & $2.005$ &  $0.76$    & $1.71$               \\ \bottomrule
		\end{tabular}
	}
	\vspace{0.1in}
	\label{tab.goal&path}
\end{table}

\begin{table}[t]
	\centering
	\caption{Quantitative comparisons between different layer number and embedding shape configurations. The default settings we finally chose are marked in  \colorbox{baselinecolor}{gray}.} 
	\scalebox{0.78}
	{
		\begin{tabular}{c|cccc|c}
			\toprule
			\multicolumn{1}{c|}{Configuration Setups} & \multicolumn{1}{c}{\begin{tabular}[c]{@{}c@{}}Goal\\ Decoder\end{tabular}} & \multicolumn{1}{c}{\begin{tabular}[c]{@{}c@{}}Path\\Decoder\end{tabular}} & \multicolumn{1}{c}{\begin{tabular}[c]{@{}c@{}}Pose\\Decoder\end{tabular}} & \multicolumn{1}{c|}{\begin{tabular}[c]{@{}c@{}}Interaction\\ Analyzer\end{tabular}} & \multicolumn{1}{c}{FID $\downarrow$} \\ \hline
			\multirow{7}{*}{Layer Number}    
			& \baseline{$4$}     & \baseline{$4$}   & \baseline{$4$} & \baseline{4} &$2.005$ \\
			& $2$     & $4$   & $4$ & $4$   & $2.037$ \\  
			& $8$     & $4$   & $4$ & $4$ &$2.006$     \\ 
			& $4$     & $2$   & $4$ & $4$ &$2.134$ \\
			& $4$     & $8$   & $4$ & $4$ &$2.005$ \\ 
			& $4$     & $4$   & $2$ & $4$ &$2.185$  \\
			& $4$     & $4$   & $8$ & $4$ &$2.004$ \\ 
			& $4$     & $4$   & $4$ & $2$ &$2.018$  \\
			& $4$     & $4$   & $4$ & $8$ &$2.009$  \\\hline
			\multirow{7}{*}{Embedding Shape}
			& \baseline{$512$}  & \baseline{$512$} & \baseline{$512$} & \baseline{$512$} & $2.005$  \\
			& $256$  & $512$   & $512$ & $512$ & $2.057$   \\
			& $1024$  & $512$   & $512$  & $512$ & $2.006$  \\ 
			& $512$  & $256$   & $512$  &  $512$ & $2.069$   \\
			& $512$  & $1024$   & $512$  & $512$ & $2.006$  \\  
			& $512$  & $512$   & $256$  & $512$ & $2.096$  \\
			& $512$  & $512$   & $1024$  & $512$ & $2.004$  \\    
			& $512$  & $512$   & $512$  & $256$ & $2.071$  \\
			& $512$  & $512$   & $512$  & $1024$ & $2.006$  \\ 
			\bottomrule     
		\end{tabular}%
	}
	\vspace{0.1in}
	\label{tab:appendix_tab1}
\end{table}

\subsubsection{Effect of Goal and Path Inferences}
In this section, we investigate the effects of GoalDecoder and PathDecoder with quantitative and qualitative analysis. Firstly, as shown in Fig. \ref{VIS_5}, we visualize inferred goal and path results to compare with their ground-truths. We can see that given a textual description as conditional input, the goal inferred from GoalDecoder is close to the real movement destination inside a 3D scene. Besides, as a non-deterministic generator, PathDecoder also performs diverse path inference based on the given textual description and inferred motion goal. Furthermore, we also report the effects of GoalDecoder and PathDecoder on the text-to-motion generation in 3D scenes. As shown in Tab. \ref{tab.goal&path}, without the inferences of motion goal and path, the performance of text-to-motion synthesis is significantly degraded, 25\% on FID and 48\% on goal error. All these quantitative and qualitative analyses verify that the proposed cascaded three-stage generation strategy significantly improves text-to-motion synthesis in 3D scenes.

\begin{figure}[t]
	\centering
	\includegraphics[width=0.95\linewidth]{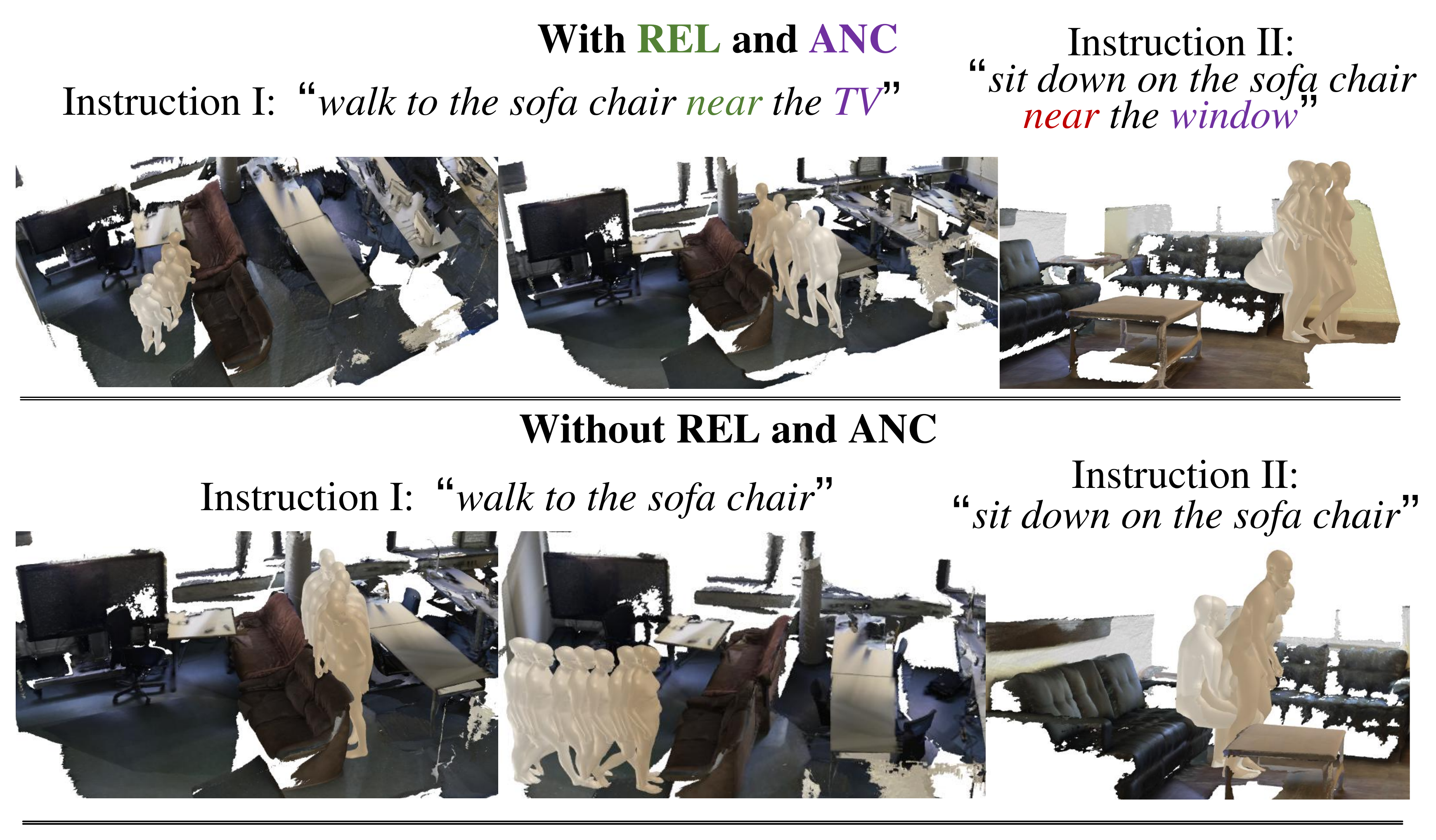}
	\vspace{-12pt}
	\caption{Human motion samples synthesized from different instruction compositions (with or without REL and ANC compositions).}
	\label{w/o_ANC_REL}
\end{figure}
\subsubsection{Effect of Given Duration}
\begin{figure}[t]
	\centering
	\includegraphics[width=0.95\linewidth]{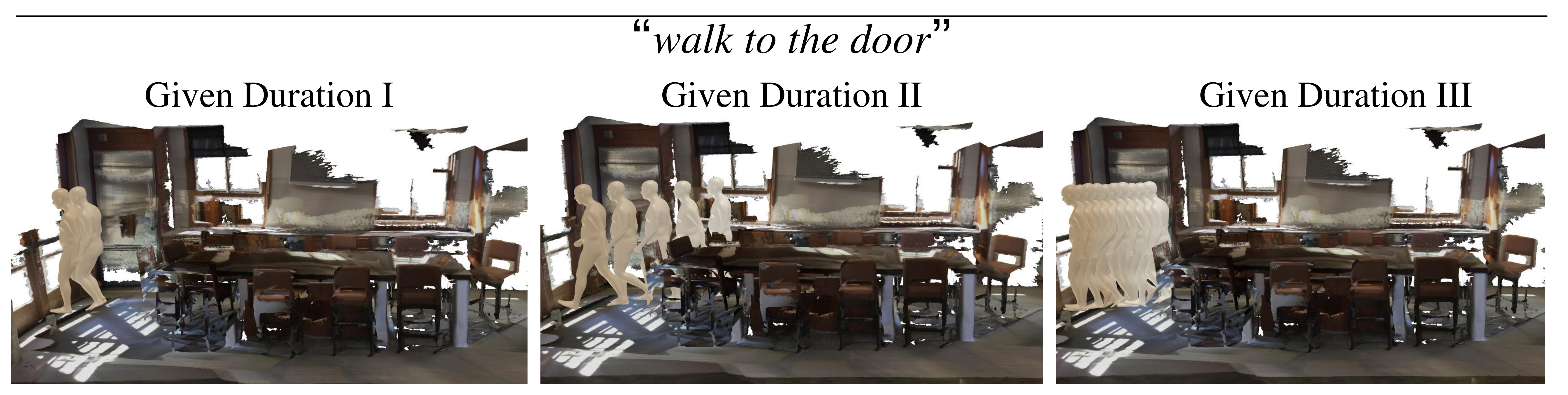}
	\vspace{-12pt}
	\caption{Human motion samples synthesized from different given durations.}
	\label{durations}
\end{figure}

\revise{To explore the effect of given duration $N$ on motion synthesis, we specify different durations in generating human motions from the same textual instruction. As shown in Fig. \ref{durations}, we respectively visualize the human motion samples synthesized from three given durations ($i.e.$, 2 frames, 5 frames, 10 frames). Although the sequence lengths of these human motion samples are different, these synthesized motions are both realistic and semantically consistent with the given textual instruction. These analyses verify that CESA is a powerful multi-modality inference system that can jointly understand 3D scene, 3D motion, textual instructions.}

\subsubsection{Effect of REL and ANC Compositions}
\revise{In this section, we analyze the effect of REL and ANC compositions in textual instructions on indoor motion syntheses. As shown in Fig. \ref{w/o_ANC_REL}, we can see that REL and ANC compositions in textual instructions characterize richer spatial layout details of the object to be interacted with. Thus, when there is more than one object of the same category in a given indoor scene, REL and ANC descriptors significantly eliminate the ambiguity in specifying objects and benefit user intention expression. These qualitative analyses verify the effectiveness of the compositional template ($i.e.$, Sr3D \cite{achlioptas2020referit3d}) we adopted in our textual descriptions.}

\subsubsection{Effect of Layer Number and Embedding Shape}
As shown in Tab. \ref{tab:appendix_tab1}, we first tune the layer of goal decoder, path decoder, pose decoder and interaction analyzer from 2, 4 to 8. We can see that deploying more layers tends to improve performance on FID. However, considering the computational cost, the performance gains brought by an over-large model are limited. Therefore, we choose 4-layer goal/path/pose decoders and 4-layer interaction analyzer as our final model configurations. Furthermore, we also investigate the configuration of their embedding shapes and prepare three size choices for them: 256, 512, and 1024 dimensions. Table \ref{tab:appendix_tab1} shows that our model is insensitive to the configuration of latent embedding shape, and the proposed synergistic synthesis-analysis strategy and cascaded text-to-motion scheme are the core reasons for observed performance improvements. 

\begin{figure}[t]
	\centering
	\includegraphics[width=0.9\linewidth]{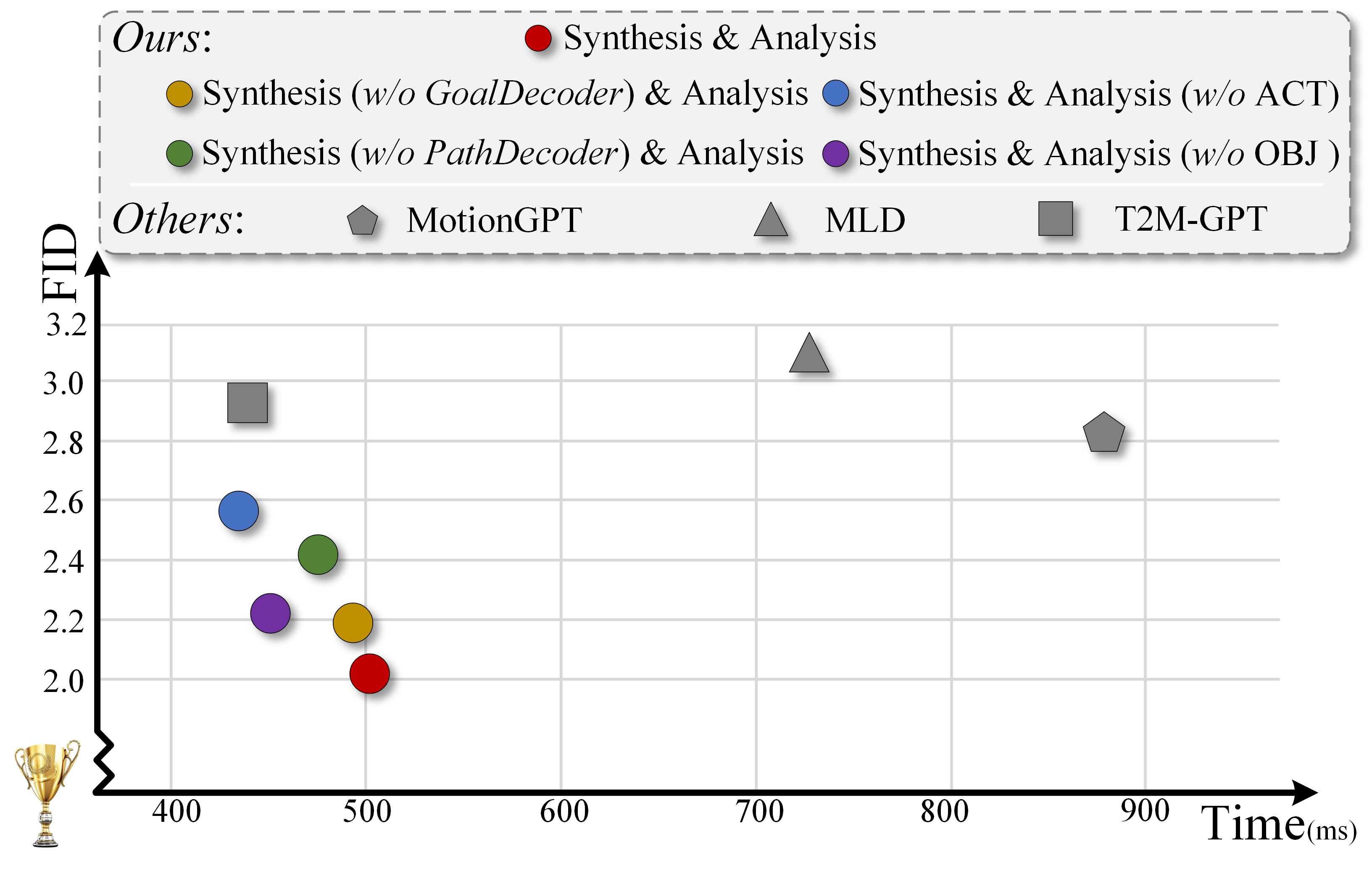}
	\vspace{-12pt}
	\caption{Comparisons between our different model configurations and other methods in terms of their FID and inference time performances. The time performance we reported is the average inference time (millisecond) of each sentence.}
	\label{infer_time}
\end{figure}
\subsection{Evaluation on Inference Time}
In this section, we compare the computational performance of CESA and other methods. As shown in Fig. \ref{infer_time}, we can see that CESA has better FID performances with less computational cost than other methods. Furthermore, it also verifies that factorizing the text-driven scene-aware human motion synthesis into cascaded three stages brings significant motion realism performance gains and limited additional computational costs. All these experimental results indicate that CESA is a lightweight yet strong baseline for the scene-aware text-to-motion task.

\section{Conclusion}
In this paper, we introduce CESA, which integrates scene-aware human motion synthesis and analysis into a synergistic pipeline and explores reciprocal benefits between them. Furthermore, we propose a cascaded generation strategy that factorizes text-driven scene-aware human motion synthesis into three stages: goal inferring, path planning, and pose synthesis. Extensive experiments verify that coupling CESA with the powerful three-stage generation strategy significantly improves text-to-motion synthesis on its motion realism and text-motion consistency while also enhancing robust scene-aware human motion analysis.  

An interesting direction for the future work of CESA is to design a powerful body-scene contact inference module. Synthesizing realistic 3D body-scene interactions from given textual instructions begins with inferring accurate body-scene contact relations. The contact inference module focuses on predicting future contact relations between every body joint and every scene point from given textual instructions. Considering predicted body-scene contact maps as priors, this body-scene contact inference module introduces richer conditions into the scene-aware text-to-motion synthesis task, significantly benefiting motion realism and motion-text semantic consistency.  

\bibliographystyle{IEEEtran}
\bibliography{IEEEabrv,ref}
\end{document}